\documentclass[letterpaper, hidelinks, 10 pt, journal, twoside]{IEEEtran}

\IEEEoverridecommandlockouts    

\usepackage{graphicx}
\usepackage{multirow}
\usepackage{algorithm,algpseudocode}
\usepackage{amsmath}
\usepackage{amsfonts}
\usepackage{amssymb}
\usepackage{amsmath}
\usepackage{amsfonts}
\usepackage{amssymb}
\usepackage{cite}
\usepackage{url}
\usepackage{threeparttable}
\usepackage{color}
\usepackage{tablefootnote}
\usepackage{tikz}
\usepackage{subcaption}
\usepackage{soul}
\usepackage{amsmath}
\usepackage{amssymb}
\usepackage{amsmath}
\usepackage{hyperref}
\usepackage{xspace}
\usepackage{tablefootnote}
\usepackage{booktabs}
\usepackage{cancel}
\usepackage[normalem]{ulem}

\newcommand{\matteo}[1]{{\color{black} #1}}

\newcommand{\morrell}[1]{{\color{black} #1}}

\newcommand{\etal}{\emph{et~al.}}
\newcommand{\mbf}[1]{\mathbf{#1}}
\newcommand{\ikdtree}{ikd-tree\xspace}
\newcommand{\BIGIkdtree}{Ikd-tree\xspace}
\newcommand{\kdtree}{kd-tree\xspace}

\newcommand{\figref}[1]{Fig.~\ref{#1}}
\newcommand{\tabref}[1]{Tab.~\ref{#1}}
\newcommand{\lidar}{lidar\xspace}
\newcommand{\lidars}{lidars\xspace}

\newcommand{\secref}[1]{Sec.~\ref{#1}}
\newcommand{\norm}[1]{\left\lVert#1\right\rVert}

\begin{document}

\title{\LARGE \bf LOCUS 2.0: Robust and Computationally Efficient Lidar Odometry for Real-Time 3D Mapping}
\author{Andrzej Reinke$^{1,2}$, Matteo Palieri$^{1,3}$, Benjamin Morrell$^{1}$, Yun Chang$^{4}$, Kamak Ebadi$^{1}$,\\ Luca Carlone$^{4}$, Ali-akbar Agha-mohammadi$^{1}$
\vspace{-1em}
\thanks{Manuscript received: February, 24, 2022; Revised May, 13, 2022; Accepted May, 19, 2022.}
\thanks{This paper was recommended for publication by Editor Javier Civera upon evaluation of the Associate Editor and Reviewers’ comments.}
\thanks{This work was supported by the Jet Propulsion Laboratory - California Institute of Technology, under a contract with the National Aeronautics and Space Administration (80NM0018D0004). This work was partially funded by the Defense Advanced Research Projects Agency (DARPA). \textcopyright  2022 All rights reserved.}
\thanks{$^{1}$Reinke, Palieri, Morrell, Ebadi and Agha-mohammadi are with NASA Jet Propulsion Laboratory, California Institute of Technology, Pasadena, CA, USA {\tt\footnotesize benjamin.morrell@jpl.nasa.gov}}
\thanks{$^{2}$Reinke is with University of Bonn, Germany {\tt\footnotesize arein@uni-bonn.de}}
\thanks{$^{3}$Palieri is with the Department of Electrical And Information Engineering, Polytechnic University of Bari, IT {\tt\footnotesize matteo.palieri@poliba.it}}
 \thanks{$^{4}$Chang and Carlone are with the Department of Aeronautics and Astronautics, Massachusetts Institute of Technology, Cambridge, MA, USA. {\tt\footnotesize lcarlone@mit.edu}}
 \thanks{Digital Object Identifier (DOI): see top of this page.}
 }
\markboth{IEEE Robotics and Automation Letters. Preprint Version. Accepted JUNE, 2022}
{Reinke \MakeLowercase{\textit{et al.}}: LOCUS 2.0} 
\maketitle
\begin{abstract}
Lidar odometry has attracted considerable attention as a robust localization method for autonomous robots operating in complex GNSS-denied environments. However, achieving reliable and efficient performance on heterogeneous platforms in large-scale environments remains an open challenge due to the limitations of onboard computation and memory resources needed for autonomous operation. In this work, we present LOCUS 2.0, a robust and computationally-efficient \lidar odometry system for real-time underground 3D mapping. 
LOCUS 2.0 includes a novel normals-based \morrell{Generalized Iterative Closest Point (GICP)} formulation that reduces the computation time of point cloud alignment, an adaptive voxel grid filter that maintains the desired computation load regardless of the environment's geometry, and a sliding-window map approach that bounds the memory consumption.
The proposed approach is shown to be suitable to be deployed on heterogeneous robotic platforms involved in large-scale explorations under severe computation and memory constraints. 
We demonstrate LOCUS 2.0, a key element of the CoSTAR team's entry in the DARPA Subterranean Challenge, across various underground scenarios.

We release LOCUS 2.0 as an open-source library and also release a \lidar-based odometry dataset in challenging and large-scale underground environments. The dataset features legged and wheeled platforms in multiple environments including fog, dust, darkness, and geometrically degenerate surroundings with a total of $11~h$ of operations and $16~km$ of distance traveled. 
\end{abstract}


\IEEEpeerreviewmaketitle

\section{Introduction}

\IEEEPARstart{L}{idar} odometry has emerged as a key tool for robust localization of autonomous robots operating in complex GNSS-denied environments. Lidar sensors do not rely on external light sources and provide accurate long-range 3D measurements by emitting pulsed light waves to estimate the range to surrounding obstacles, through time-of-flight-based techniques. For these reasons, \lidar has been often preferred over visual sensors to achieve reliable ego-motion estimation in cluttered environments with significant illumination variations (e.g., search, rescue, industrial inspection and underground exploration). 

Lidar odometry algorithms aim to recover the robot's motion between consecutive \lidar acquisitions using scan registration. Through repeated observations of fixed environmental features, the robot can simultaneously estimate its movement, construct a map of the unknown environment, and use this map to keep track of its position within it. 

\begin{figure}[t!]
\centering
	\includegraphics[width=1\columnwidth, trim= 0mm 0mm 0mm 0mm, clip]{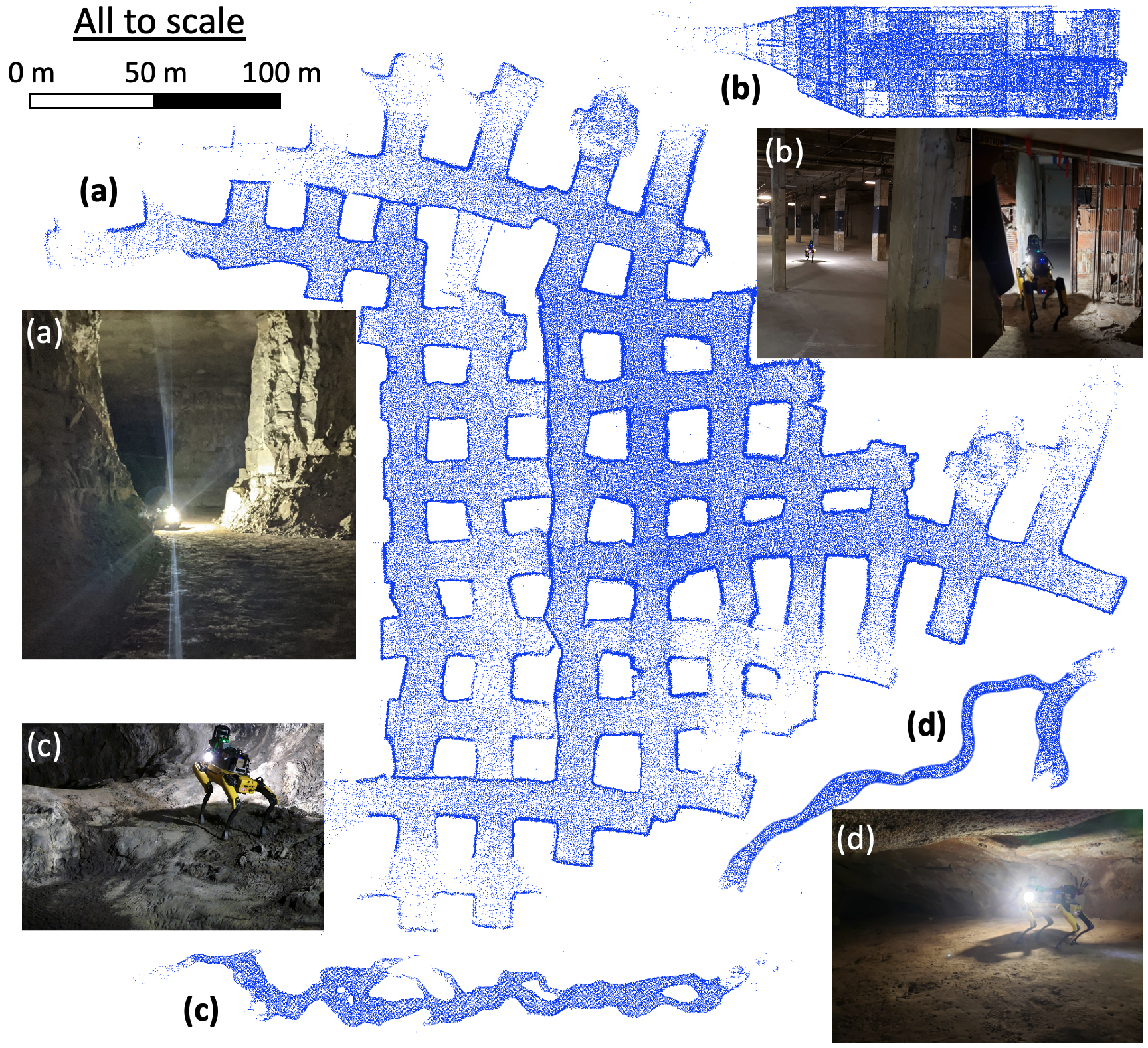}
	\caption{\footnotesize{Four examples from our large underground \lidar-based SLAM dataset consisting of over $16~km$ distance traveled and $11~h$ of operation across diverse environments: (a) a large-scale Limestone Mine (Kentucky Underground), (b) A 3-level urban environment with both large, open spaces and tight passages (LA Subway), (c) lava tubes with large vertical changes, and (d) lava tubes with narrow passages. LOCUS 2.0 performed successfully in all these environments on computationally constrained robots.}}
	\label{fig:locus_spot1_final}
\end{figure}

While many \lidar odometry algorithms can achieve remarkable accuracy, their computational cost can still be prohibitive for computationally-constrained platforms, reducing their field of applicability in systems of heterogeneous robots, where some of the robots may have very limited computational resources. Moreover, many existing approaches maintain the global map in memory for localization purposes, making them unsuitable for large-scale explorations where the map size in memory would significantly increase. 

Our previous work \cite{palieri2020locus}
presents LOCUS 1.0, a multi-sensor \lidar-centric solution for high-precision odometry and 3D mapping in real-time featuring a multi-stage scan matching module, equipped with health-aware sensor integration that fuses additional sensing modalities in a loosely-coupled scheme. While achieving remarkable accuracy and robustness in perceptually degraded settings, the previous version of LOCUS 1.0: \textit{i)} had a more significant computational load, \textit{ii)} maintained the global map in memory, \textit{iii)} was less robust to more generalized sensor failures, e.g., failure of one of \lidar sensor. 
LOCUS 2.0 presents algorithmic and system-level improvements to decrease the computational load and memory demand, enabling the system to achieve accurate and real-time ego-motion estimation in challenging perceptual conditions over large-scale exploration under severe computation and memory constraints. 

The new features and contributions of this work include ~(i)~\textit{GICP from normals}: a novel formulation of \morrell{Generalized Iterative Closest-Point (GICP)} that leverages point cloud normals to approximate the point covariance calculation for enhanced computational efficiency,~(ii) \textit{Adaptive voxel grid filter} that ensures deterministic and near-constant runtime, independently of the surrounding environment and \lidars,~(iii) improvement and evaluation of two \textit{sliding-window map} storage data structures: multi-threaded octree, \ikdtree~\cite{cai2021ikd}, and~(iv) dataset release{\footnote{\url{https://github.com/NeBula-Autonomy/nebula-odometry-dataset}}} including in challenging, real-world subterranean environments (urban, tunnel, cave), shown in \figref{fig:locus_spot1_final}, collected by heterogeneous robot platforms.
All these features improve the computational and memory operation while maintaining accuracy at the same level.
The source code of LOCUS 2.0 has been released as an open-source library{\footnote{\url{https://github.com/NeBula-Autonomy/LOCUS}}}.

The paper is organized as follows. \secref{sec:related_work} reviews related work on \lidar odometry. \secref{sec:system_description} describes the proposed system architecture with a focus on the updates made to the framework to enhance its performance in terms of computational load and memory usage. \secref{sec:experiments} provides an ablation study of the system
on datasets collected by heterogeneous robots during the three circuits of the DARPA Subterranean Challenges, an international robotic competition where robots are tasked to explore complex GNSS-denied underground environments autonomously.

\section{Related Works}
\label{sec:related_work}

Motivated by the need to enable real-time operation under computation constraints and large-scale explorations under memory constraints in perceptually-degraded settings, we review the current state-of-the-art to assess whether any solution can satisfy these requirements simultaneously.

\subsection{Lidar Odometry} The work \cite{han2021dilo} proposes DILO, a \lidar odometry technique that projects a three-dimensional point cloud onto a two-dimensional spherical image plane and exploits image-based odometry techniques to recover the robot ego-motion in a frame-to-frame fashion without requiring map generation. This results in dramatic speed improvements, however, the method does not fuse additional sensing modalities and is not open-source. 
The work \cite{liu2021balm} presents BALM, a \lidar odometry solution exploiting bundle adjustment over a sliding window of \lidar scans for enhanced mapping accuracy. While the paper claims nearly real-time operation, the method does not fuse sensing modalities and maintains the entire map in memory. 

\subsection{Lidar-Inertial Odometry} General challenges encountered in pure \lidar-based odometry estimators include degraded motion estimation in high-rate motion scenarios \cite{zhang2014loam}, and degenerate motion observability in geometrically-featureless areas (e.g. long corridors, tunnels) \cite{tagliabue2021lion,ebadi2021dareslam}. For these reasons, \lidars are commonly fused with additional sensing modalities to achieve enhanced accuracy in perceptually-degraded settings \cite{shan2021lvi, zhao2021super}.
The work \cite{shan2020lio} presents LIO-SAM, an accurate tightly-coupled \lidar-inertial odometry solution via smoothing and mapping that exploits a factor graph for joint optimization of IMU and lidar constraints. Scan-matching at a local scale instead of a global scale significantly improves the real-time performance. 
The work \cite{li2021towards} presents LILI-OM, a tightly-coupled \lidar-inertial odometry solution with a \lidar/IMU hierarchical keyframe-based sliding window optimization back-end. 
The work \cite{qin2020lins} presents LINS, a fast tightly-coupled fusion scheme of \lidar and IMU
with error-state Kalman filter to recursively correct the estimated state by generating new feature correspondences in each iteration. 
The work \cite{yang2021rtlio} presents RTLIO, a tightly-coupled \lidar-inertial odometry pipeline that delivers accurate and high-frequency estimation for the feedback control of UAVs by solving a cost function consisting of \lidar and IMU residuals.
The work \cite{xu2021fast} presents FAST-LIO, a computationally-efficient \lidar-inertial odometry pipeline that fuses \lidar feature points with IMU data in a tightly-coupled scheme with an iterated extended Kalman filter. 
A novel formula for computing the Kalman gain results in a considerable decrease of computational complexity with respect to the standard formulation, translating into decreased computation time. 
The work \cite{chen2021direct} presents DLIO, a lightweight loosely-coupled \lidar-inertial odometry solution for efficient operation over constrained platforms. The work provides efficient derivation of local submaps for global refinement constructed by concatenating point clouds associated with historical key-frames, along with a custom iterative closest point solver for fast and lightweight point cloud registration with data structure recycling that eliminates redundant calculations. 
While these methods are computationally efficient, the methods maintain a global map in memory, rendering it unsuitable for large-scale explorations over memory constraints.

\subsection{Lidar-Visual-Inertial Odometry} The work \cite{zhao2021super} presents Super Odometry, a robust, IMU-centric multi-sensor fusion framework that achieves accurate operation in perceptually-degraded environments. The approach divides the sensor data processing into several sub-factor-graphs where each sub-factor-graph receives the prediction from an IMU pre-integration factor, recovering the motion from a coarse to fine manner and enhancing the real-time performance. The approach also adopts a dynamic octree data structure to organize the 3D points, making the scan-matching process very efficient and reducing the overall computational demand. However, the method maintains the global map in memory.

The work \cite{shan2021lvi} presents LVI-SAM, a real-time tightly-coupled \lidar-visual-inertial odometry solution via smoothing and mapping built atop a factor graph, comprising a visual-inertial subsystem (VIS) and a \lidar-inertial subsystem (LIS).
However, the method maintains the global map in memory, which it not unsuitable for large-scale explorations with memory limited processing units. 
The work \cite{lin2021r2live} proposes R2LIVE, an accurate and computationally-efficient sensor fusion framework for \lidar, camera, and IMU that exhibits extreme robustness to various failures of individual sensing modalities through filter-based odometry.
While not explicitly mentioned in the paper, the open-source implementation of this method features the integration of an \ikdtree data structure for map storage which could be exploited to keep in memory only a robot-centered submap. 

\section{System Description}
\label{sec:system_description}

\begin{figure}
\vspace{0.5em}
	\centering     
	\includegraphics[width = 1.0\linewidth]{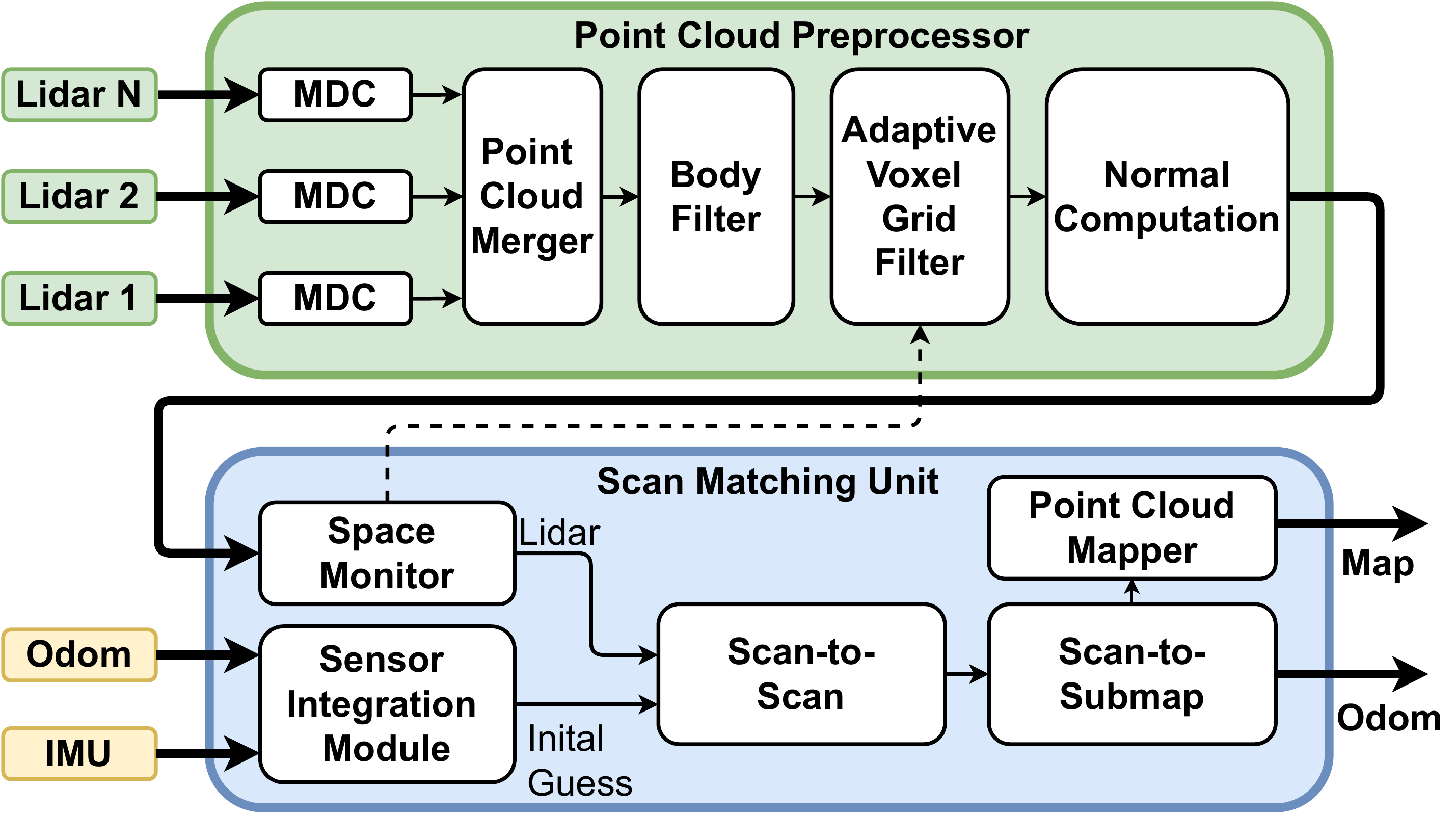}
	\caption{\footnotesize{LOCUS 2.0 architecture.
	 }}
	\label{fig:new_locus}
\end{figure}

LOCUS 2.0 provides an accurate Generalized Iterative Closest Point (GICP) algorithm \cite{GICP} based multi-stage scan matching unit and a health-aware sensor integration module for robust fusion of additional sensing modalities in a loosely coupled scheme.
The architecture, shown in \figref{fig:new_locus}, contains three main components:
i) \textit{point cloud preprocessor}, ii) \textit{scan matching unit}, iii) \textit{sensor integration module}.
The \textit{point cloud preprocessor} is responsible for the management of multiple-input \lidar streams to produce a unified 3D data product that can be efficiently
processed by the \textit{scan matching unit}.
The \textit{preprocessor module} consists of Motion Distortion Correction (MDC) of the point clouds. 
This module corrects the distortion in the point cloud from sensor rotation during a scan due to robot movement using IMU measurements.

\matteo{Next, the \textit{Point Cloud Merger} enlarges the robot field-of-view by combining point clouds from different \lidar sensors in the robot body frame using their known extrinsic transformation. To enable resilient merging of multiple \lidar feeds, we introduce an external timeout-based health monitor that dynamically updates which \lidars should be combined in the \textit{Point Cloud Merger} (i.e. a \lidar is ignored if its messages are too delayed). The health monitoring makes the submodule robust to lags and failures of individual \lidars so that an output product is always provided to the downstream pipeline.} 
Then, the \textit{Body Filter} removes the 3D points that belong to the robot. Next, the \textit{Adaptive Voxel Grid Filter} maintains a fixed number of voxelized points to manage CPU load and to ensure deterministic behavior. 
It allows the robot to have consistent computational load regardless of the size of the environment or the number of \lidars (or if potential \lidar failures).
In comparison to LOCUS 1.0, the \textit{Adaptive Voxel Grid Filter} changes the strategy of point cloud reduction from a blind voxelization strategy with fixed leaf size and random filter to an adaptive system (\secref{sec:theory:adaptive_voxelization}).
The \textit{Normal Computation} module calculates normals from the voxelized point cloud. The \textit{scan matching unit} performs a GICP scan-to-scan and scan-to-submap registration to estimate the 6-DOF motion of the robot.
LOCUS 2.0, in comparison to its predecessor, does not recalculate covariances but instead leverages a novel GICP formulation to use normals, which only need to be computed once and stored in the map (\secref{sec:theory:gicp_from_normals}). 

In robots with multi-modal sensing, when available, LOCUS 2.0 uses an initial  estimate from a non-\lidar source (from Sensor Integration Module) to ease the convergence of the GICP in the scan-to-scan matching stage, by initializing the optimization with a near-optimal seed that improves accuracy and reduces computation, enhancing real-time performance, as explained in \cite{palieri2020locus}. 

LOCUS 2.0 also includes a more efficient technique for map storage.
The system uses a sliding-window approach because large-scale areas are not feasible to be maintained in memory.
For example, in one of the cave datasets presented here, a global map at $1~cm$ resolution requires $50~GB$ of memory, far exceeding the typically available memory on small mobile robots. 
This approach demands efficient computational solutions for insertion, deletion, and search.

\subsection{\textit{GICP from normals}}
\label{sec:theory:gicp_from_normals}
LOCUS 2.0 uses GICP for scan-to-scan and scan-to-submap matching.
GICP generalizes the point-to-point and point-to-plane ICP registration by using a probabilistic model for the registration problem \cite{GICP}.
To do this, GICP requires the availability of covariances for each point in the point clouds to be aligned. 
Covariances are usually calculated based on the distributions of neighboring points around a given point.
Segal \etal \cite{GICP} presents plane-to-plane application with the assumption that real-world surfaces are at least locally planar.
In this formulation, points on surfaces are locally represented by a covariance matrix, where the point is known to belong to a plane with high-confidence, but its exact location in the plane has higher uncertainty.

Here, we show how plane-to-plane covariance calculation is equivalent to calculating covariances from pre-computed normals.
The fact that only the normal is needed is especially important for scan-to-submap alignment since the map would otherwise require recomputing point covariances, which is an expensive operation involving the creation of a \kdtree~and nearest neighbors search. By instead using normals, the covariance computation is only performed once (since it is not influenced by the addition of extra points), and the result can be stored and reused.

\matteo{The most common implementations of GICP~\cite{rusu20113d},~\cite{Koide2020} rely on computing the covariances $C_i^B$ and $C_i^A$ for each point $i$ in two scans.
\morrell{That calculation of covariances in GICP takes place as a pre-processing step whenever two scans have to be registered. In the following, we describe how to obtain the covariances, without recomputing them every time a scan is collected.}
Any well-defined covariance matrix $\mbf{C}$ can be eigendecomposed to eigenvalues and eigenvectors \cite{strang09}, $ \mbf{C}  = \lambda_1~\boldsymbol{u_1} \cdot \boldsymbol{u_1}^T +  \lambda_2~\boldsymbol{u_2}  \cdot  \boldsymbol{u_2}^T +  \lambda_3~\boldsymbol{u_3} \cdot  \boldsymbol{u_3}^T $
where $\lambda_1$, $\lambda_2$, $\lambda_3$ are the eigenvalues of matrix $\mbf{C}$, and $\boldsymbol{u_1}$, $\boldsymbol{u_2}$, $ \boldsymbol{u_3}$ are eigenvectors of matrix $\mbf{C}$.
Two eigenvalues with the same value can be interpreted and visualized as eigenvectors that span equally 2D planar surface with the same distribution in each direction. This allows the covariance computation problem to be thought of geometrically.

In the plane-to-plane metric, for a point $a_i$ from scan $A$ we know the position along the normal with very high confidence, but we are less sure about its location in the plane. 
 \morrell{To represent this, we set $\boldsymbol{u_1}=\boldsymbol{n}$, and assign $\lambda_1=\epsilon$ as the variance in the normal direction, where $\epsilon$ is a small constant. We can then choose the other eigenvectors to be arbitrary vectors in the plane orthogonal to $\boldsymbol{n}$ and assign them comparatively large eigenvalues $\lambda_2=\lambda_3=1$, indicating that we are unsure about location in the plane.}
 
Let us take any vector that lies on the plane and is perpendicular to the normal.
The vector needs to satisfy the plane equation (that it is perpendicular to the normal vector) and cross the origin:
$n_xx + n_yy+n_zz = 0$.
Then $z~=~-\frac{n_x \cdot x + n_y \cdot y }{n_z}$, therefore the family of vectors on the plane is
$
	\boldsymbol{u_2} = \frac{(x, y, -(n_x \cdot x + n_y \cdot y)/ n_z)}{ \norm{(x, y, -(n_x \cdot x + n_y \cdot y) / n_z}}
$,
where $n_z$ and $n_x$ corresponds to the component $z$ and $x$ of a normal vector $\textbf{n}$ and $n_z=0$ means the horizontal vector.
The third vector $\boldsymbol{u_3}$ needs to simultaneously be perpendicular to $\boldsymbol{u_1}$ and $\boldsymbol{u_2}$ since eigenvectors need to span the whole $3D$ space. Therefore, $\boldsymbol{u_3} = \boldsymbol{n} \times \boldsymbol{u_2}$.
If we know the eigenvectors and eigenvalues of matrix $\mbf{C}$, we have
{$
	\mbf{C}  = \epsilon~\boldsymbol{u_1}  \cdot \boldsymbol{u_1}^T +  1.0~\boldsymbol{u_2}   \cdot  \boldsymbol{u_2}^T +  1.0~\boldsymbol{u_3}   \cdot  \boldsymbol{u_3}^T 
$}. Substituting from above, we get:
\begin{equation}
\begin{aligned}
	\mbf{C}  = &  \epsilon~\boldsymbol{n}  \cdot \boldsymbol{n}^T  
	+ 1.0~\boldsymbol{u_2}   \cdot  \boldsymbol{u_2}^T  
	+ 1.0~(\boldsymbol{n} \times \boldsymbol{u_2}) \cdot  (\boldsymbol{n} \times \boldsymbol{u_2}^T)
\end{aligned}
\end{equation}
Then, if we take arbitrarily $x=1$ and $y=0$ then $z=-\frac{n_x}{n_z}$, and the covariance simplifies to:
\begin{align}
	\mbf{C}  = & \epsilon~\boldsymbol{n}  \cdot \boldsymbol{n}^T  +  \frac{(1, 0, -n_x/n_z)}{\norm{(1, 0, -n_x/n_z)}}  \cdot  \frac{(1, 0, -n_x/n_z)}{\norm{(1, 0, -n_x/n_z)}}^T  \\
	+  &  \boldsymbol{n} \times \frac{(1, 0, -n_x/n_z)}{\norm{(1, 0, -n_x/n_z)}} \cdot 
	 \boldsymbol{n} \times \frac{(1, 0, -n_x/n_z)}{\norm{(1, 0, -n_x/n_z)}}^T \nonumber
	\label{eqn:final}
\end{align}
The results mean that the covariance can be purely expressed via precomputed normals at each point.
}
\subsection{Adaptive Voxel Grid Filter}
\label{sec:theory:adaptive_voxelization}
To manage the computation load of \lidar odometry, regardless of the environment
and \lidar configuration (in terms of number of \lidars and types), we propose an adaptive voxel grid filter.
In this approach, the goal is to maintain the voxelized number of points at a fixed level (desired by the user) rather than specifying the voxel leaf size and exposing the system to the variability of the input points that stems from different sensors configurations and cross-sectional geometry of the environment.
This design goal comes from the fact that almost all computations in the registration stages are dependent on a number of points $N$.
Therefore the idea is to keep the voxelized number of $3D$ points fixed to have approximately fixed computation time per scan. 
The approach is as follows: let us take any size of the initial voxel size $d_{init}$ and set $d_{leaf} = d_{init}$, where $d_{leaf}$ is the size of the voxel leaf in the current time stamp.
We propose the following control scheme:
$
d_{leaf_{t+1}}=d_{leaf_{t}}\frac{N_{scan}}{N_{desired}}
$.
The formula describes how much should the current voxel size change $d_{leaf_{t+1}}$ in comparison to what the current size is $d_{leaf_{t}}$ based on the ratio on the number of points in the current input scan $N_{scan}$ to the points that are desired for computation for given robot $N_{desired}$. \morrell{This simple technique maintains the number of points on the fixed level, while avoiding any large jumps in the numbers of points, having too few points (e.g. a faulty scan) or having too many points. The result is an improvement in the efficiency and reduction of the computational load of the system.}

\begin{figure}
\vspace{0.5em}
     \centering
     \includegraphics[width=0.8\columnwidth]{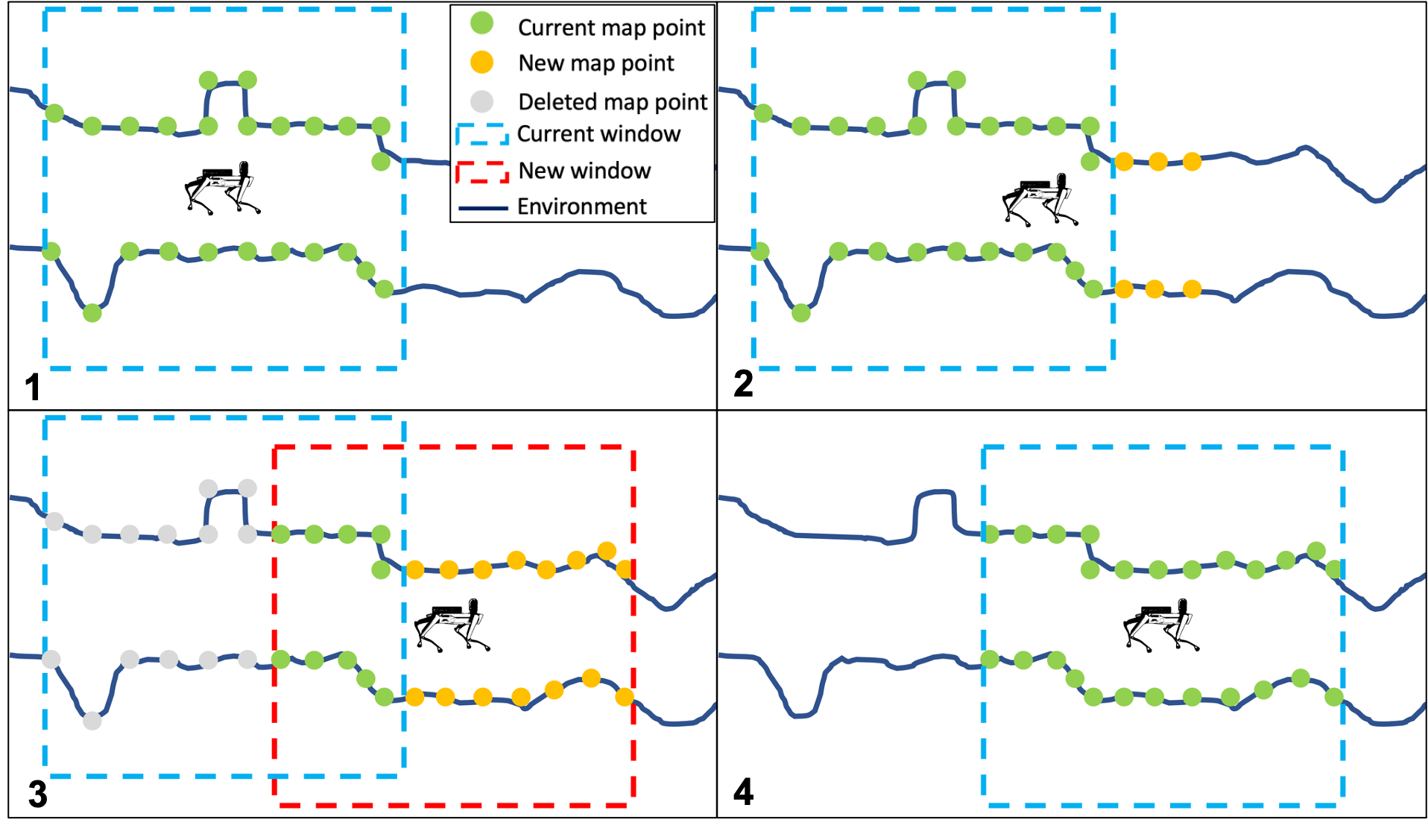}
     \caption{\morrell{\footnotesize{Illustration of our our sliding map approach. All points are maintained until the robot reaches the boundary of the original window (step 3). Then, a new window is set, and points outside that window are deleted.}}
        }        
         		\label{fig:sliding_window5}
\end{figure}

\subsection{Sliding-window Map}

LOCUS 1.0 \cite{palieri2020locus} stored the global map in memory through an octree data structure. The native octree implementation does not have an efficient way to prune data out. While a possible workaround is to filter the points around the robot and rebuild the octree accordingly, this might be computationally expensive and lead to long refreshing times.

To account for these challenges, and enable large-scale explorations under memory constraints, LOCUS 2.0 provides two map sliding-window approaches (\figref{fig:sliding_window5}): i) multi-threaded octree, ii) incremental k-dtree \cite{cai2021} (\ikdtree). 

\textbf{Multi-Threaded Octree} approach maintains only a robot-centered submap of the environment in memory. Two parallel threads ($thread_a$ and $thread_b$) each working on dedicated data structures ($map_a$/$octree_a$ and $map_b$/$octree_b$) are responsible to dynamically filter the point cloud map around the current robot position through a box-filter, and rebuild the octree accordingly with the updated map, while accounting for robot motions between parallel worker processes.

\textbf{\BIGIkdtree} \cite{cai2021} is a binary search tree that dynamically stores $3D$ points by merging new scans. 
\BIGIkdtree does not maintain 3D points only in the leaf nodes: they have points in the internal nodes as well.
This structure allows dynamic insertion and deletion capabilities and relies on lazy labels storage across the whole data structure.
Initial building of an~\ikdtree~is 
similar to a ~\kdtree, where space is split at the median point along the longest dimension recursively.
Points that are moved out of the boundaries of the~\ikdtree~data structure are not deleted immediately, but they are labeled as $deleted = True$ and maintain information until a  rebalancing procedure is triggered.

\section{Experimental Results} 
\label{sec:experiments}
\subsection{Dataset}
\begin{figure}[]
\vspace{0.5em}
     \centering
     \begin{subfigure}[]{0.2\textwidth}
         \centering
         \includegraphics[width=\textwidth]{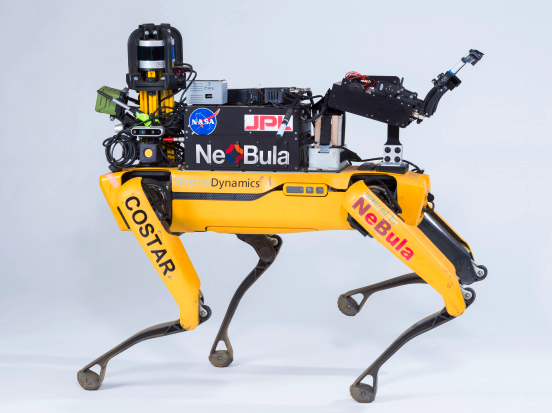}
         \caption{\footnotesize{NeBula Spot robot}}
         \label{fig:spot}
     \end{subfigure}\hfill
     \begin{subfigure}[]{0.2\textwidth}
         \centering
         \includegraphics[width=\textwidth]{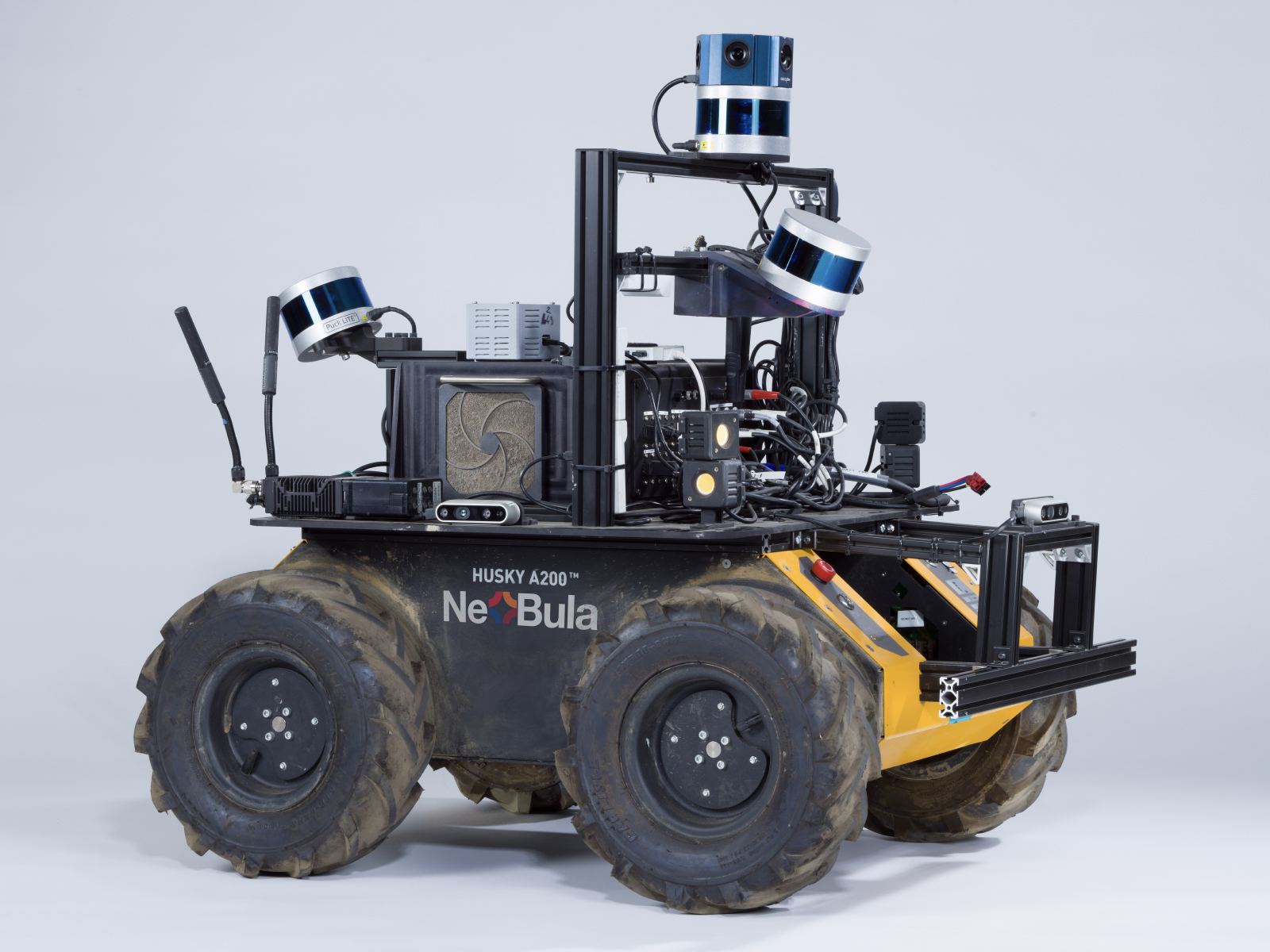}
         \caption{\footnotesize{NeBula Husky robot}}
         \label{fig:husky}
     \end{subfigure}
        \caption{\footnotesize{Type of robots for heterogeneous robotic system in Nebula framework for  DARPA Subterranean Challenge}}
       	\label{fig:spot_husky_robot}
\end{figure}

Over the last $3$ years, Team CoSTAR~\cite{nebula} has  
intensively 
tested our \lidar odometry system in real world environment such as caves, tunnels and abandoned factories. 
\begin{table}
\vspace{0.5em}
\caption{\footnotesize{Dataset summary.}}
\label{tab:tabledataset}
\begin{center}
	\resizebox{1.0\linewidth}{!}{
	\begin{threeparttable}
	\begin{tabular}{c c c c c c c c} \hline
		ID & Place & Robot & \begin{tabular}{@{}c@{}}Distance\\ (m) \end{tabular}	& \begin{tabular}{@{}c@{}}Duration\\ (min) \end{tabular} & Characteristic & \lidars  \\ 
		\toprule
		A & \begin{tabular}{@{}c@{}}power plant\\ Elma, WA \\ (Urban)\end{tabular}   & Husky & 631.53 & 59:56 & \begin{tabular}{@{}c@{}}feature-poor corridors, \\ large open spaces\end{tabular} & 3 \\ 
		\midrule
		B & \begin{tabular}{@{}c@{}}power plant\\ Elma, WA \\ (Urban) \end{tabular}  & Spot  & 664.27  & 32:26 & \begin{tabular}{@{}c@{}}2-level, \\ stairs, \\ feature-poor corridors, \\ large \& narrow spaces\end{tabular} & 1\\
		\midrule
		C & \begin{tabular}{@{}c@{}}power plant\\ Elma, WA \\ (Urban)\end{tabular}  & Husky & 757.40 & 24:21 & \begin{tabular}{@{}c@{}}feature-poor corridors, \\ large \& narrow spaces\end{tabular} & 3\tnote{\textbf{*}} \\
		\midrule
		D & \begin{tabular}{@{}c@{}}Bruceton  Mine \\ Pittsburgh, PA \\ (Tunnel) \end{tabular}   & Husky & 1795.88 & 65:36 & \begin{tabular}{@{}c@{}}self-similar\\ self-repetitive geometries\end{tabular} & 3\tnote{\textbf{*}} \\
		\midrule
		E & \begin{tabular}{@{}c@{}}Lava Beds National \\ Monument, CA \\ (Cave)\end{tabular}   & Spot & 590.85 & 25:20 & \begin{tabular}{@{}c@{}} lava tubes and pools, \\ non-uniform environment, \\  degraded lightning\end{tabular} & 1\\
		\midrule
		F & \begin{tabular}{@{}c@{}}Bruceton  Mine \\ Pittsburgh, PA \\ (Tunnel)\end{tabular}   & Husky & 1569.73 & 49:13 & \begin{tabular}{@{}c@{}}self-similar\\ self-repetitive geometries\end{tabular} & 3\tnote{\textbf{*}}\\ 
		\midrule
		G & \begin{tabular}{@{}c@{}}power plant\\ Elma, WA \\ (Urban)\end{tabular}   & Husky & 877.21 & 93:10 & \begin{tabular}{@{}c@{}}feature-poor corridors, \\ large \& narrow spaces\end{tabular} & 3\\ 
		\midrule
		H & \begin{tabular}{@{}c@{}}Subway Station \\ Los Angeles, CA \\ (Urban)\end{tabular}   & Spot & 1777.45  & 46:57 & \begin{tabular}{@{}c@{}}3-level, \\ multiple stairs, \\ feature-poor corridors, \\ large \& narrow spaces\end{tabular} & 3 \\ 
		\midrule
		I & \begin{tabular}{@{}c@{}}Kentucky Underground\\ Limestone Mine, KY \\ (Cave)\end{tabular}   & Spot & 768.82 & 19:28 & \begin{tabular}{@{}c@{}} large area, \\ non-uniform environment, \\  degraded lightning\end{tabular} & 1 \\ 
		\midrule
		J & \begin{tabular}{@{}c@{}}Kentucky Underground\\ Limestone Mine, KY \\ (Cave)\end{tabular}   & Husky & 2339.81 & 57:55 & \begin{tabular}{@{}c@{}} large area, \\ non-uniform environment, \\  degraded lightning\end{tabular} & 3\\
		\bottomrule
	\end{tabular}
\begin{tablenotes}
\item[\textbf{*}] \matteo{For our experiments we use only two \lidars.}
\end{tablenotes}
\end{threeparttable}
}
\end{center}
\end{table}
Each dataset (\tabref{tab:tabledataset}) is selected to contain components that are challenging for \lidar odometry. 
The dataset provides \lidar scans, IMU and \matteo{wheeled inertial odometry} (WIO) measurements, as well cameras stream.
All datasets have been recorded on different robotics platforms, e.g., Husky and Spot (\figref{fig:spot_husky_robot}) with 
vibrations and large accelerations as is characteristic of both a skid-steer wheeled robot traversing rough terrain and a legged robot that slips and acts dynamically in rough terrain.
The Husky robot is equipped with $3$ on-board VLP16 \lidar sensors extrinsic calibrated (one flat, one pitched forward ~30 $\deg$, one pitched backward ~30 $\deg$). 
The Spot robot is equipped with one on-board \lidar sensor extrinsic calibrated. 
Spot out-of-the-box implements \matteo{(kinematic inertial odometry)} $KIO$ and \matteo{(visual inertial odometry)} $VIO$, therefore the data records those readouts as well.
Lidar scans are recorded at $10~Hz$. WIO and IMU are recorded at $50~Hz$.
To determine the ground truth of the robot in the environment, a survey-grade $3D$ map (provided by DARPA in the Subterranean Challenge or produced by the team) is used. 
The ground-truth trajectory is produced by running LOCUS 1.0 against the survey-grade map (i.e. scan-to-map is scan-to-survey-map). In this mode, LOCUS 1.0 is tuned for maximum accuracy at the cost of computational efficiency, as it does not need to be run in real-time.  
The ground truth trajectory of the robot is determined based on LOCUS 1.0 and its multi-stage registration technique: scan-to-scan and scan-to-map (with high computational parameters and slower pace of data processing) and some manual post-processing work. 
These datasets have been made open-source to promote further research on \lidar odometry and SLAM in underground environments: \url{github.com/NeBula-Autonomy/nebula-odometry-dataset}. 

\subsection{Metrics}
For \textit{CPU and memory profiling}, a cross-platform library for retrieving information on running processes and system utilization 
is used \cite{psutil}.
The library is used for system monitoring and profiling. 
The CPU represents the percentage value of the current system-wide CPU utilization, where 100\% means 1 core is used.
The memory represents statistics by summing different memory values depending on the platform. 
\textit{Odometry delay} measures the difference between odometry message creation and the current timestamp of the system. 
The system is implemented in Robot Operating System (ROS) framework. 
This work considers maximum delay and \matteo{mean} delay since those two metrics more directly impact the performance of the modules using the odometry result, e.g., controllers and path planners.
\textit{Lidar callback time} measures the duration time for a scan at the time stamp $t_k$ to go through a pipeline of processing 
from the queue of the \lidar scans.
\textit{Scan-to-scan time} measures the duration time for a scan at time $t_k$ to align with a scan at time $t_{k-1}$ in GICP registration stage.
\textit{Scan-to-submap time} measures the duration time for a pre-aligned scan from scan-to-scan at time $t_k$ to align with a reference local map in GICP registration.

\subsection{Computation time}

\subsubsection{GICP from normals}
\label{sec:gicp_from_normals}

The experiments presented in this section are designed to show the benefit of \textit{GICP from normals} over GICP and support the claim that this reformulation yields better computation performance without sacrificing accuracy.
For each dataset, we compute statistics over $5$ runs. 
The GICP parameters for this experiment are chosen based on \cite{Reinke2022masterthesis}.
The parameters for scan-to-scan and scan-to-submap are the same: optimization step $1\mathrm{e}{-10}$, maximum corresponding distances for associations $0.3$, maximum number of iterations in optimization $20$, rotational fitness score threshold $0.005$.
Husky computation runs in $4$ threads, while Spot uses only $1$ thread due to CPU limitations.
The octree stores the map with a leaf size $0.001~m$.

\begin{figure*}[]
\vspace{0.2cm}
    \centering
    \begin{subfigure}{0.32\textwidth}
      \centering
      \includegraphics[width=1.0\textwidth]{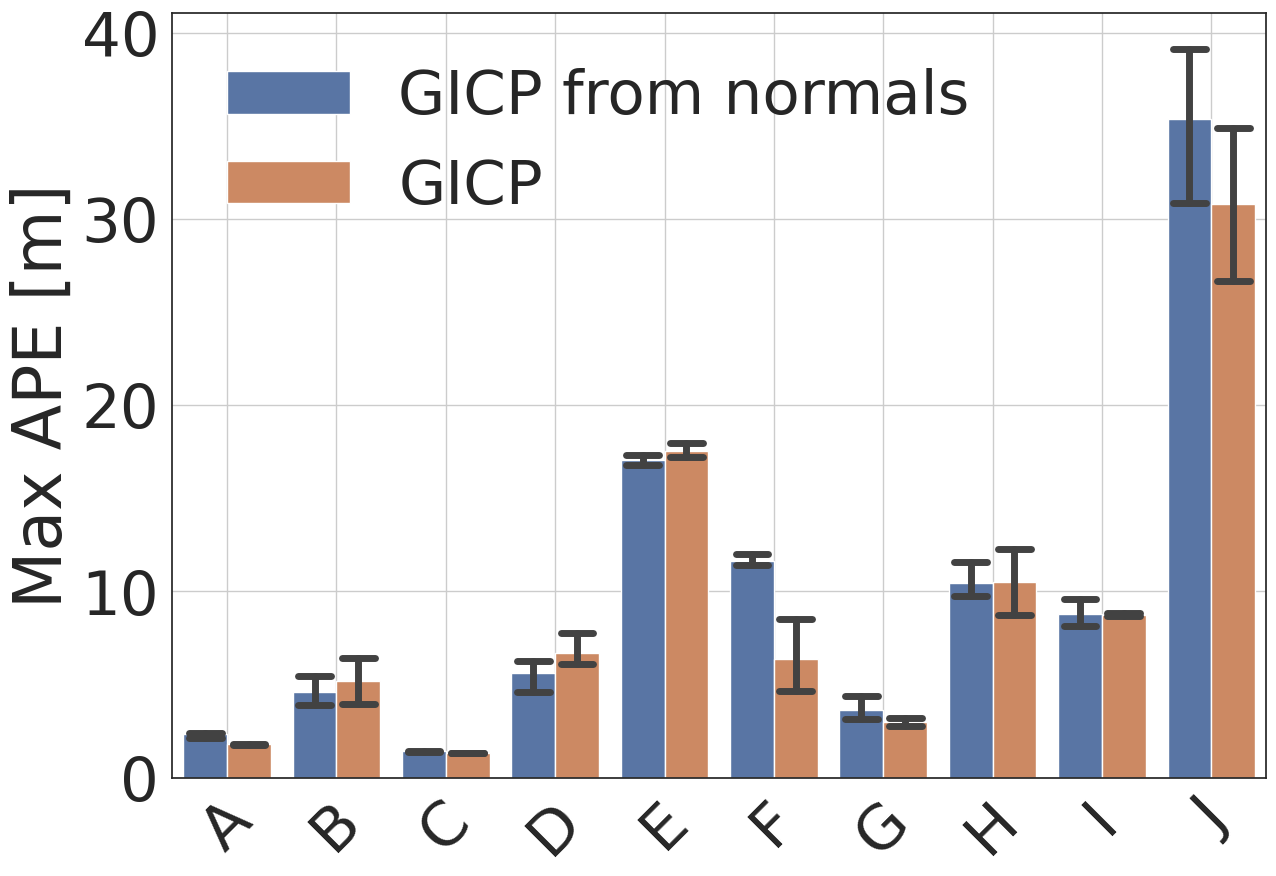}
      \label{fig:exp:max_ape}
    \end{subfigure}
    \begin{subfigure}{0.32\textwidth}
      \centering
      \includegraphics[width=1.0\textwidth]{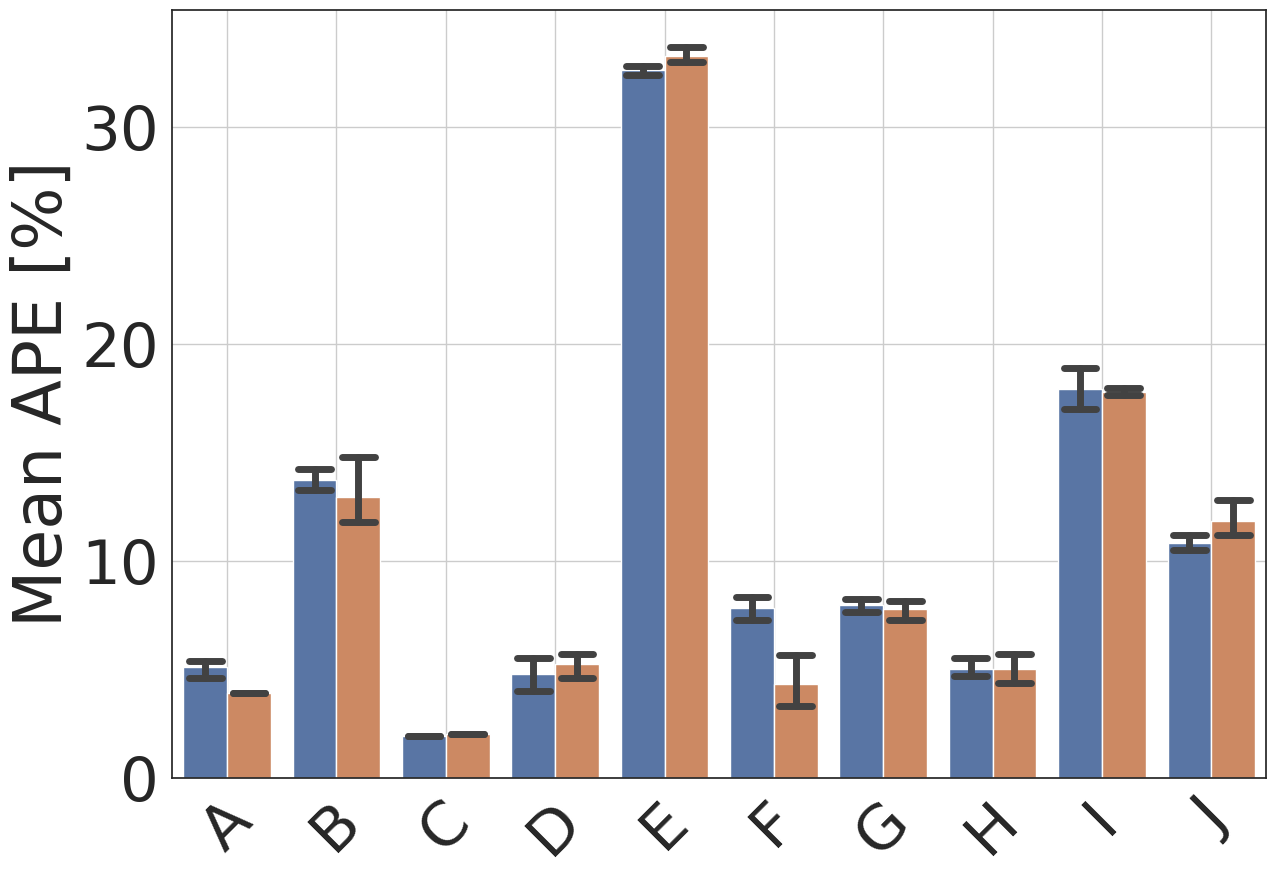}
      \label{fig:exp:mean_ape}
    \end{subfigure}
\vspace{-0.5cm}
        \begin{subfigure}{0.32\textwidth}
      \centering
      \includegraphics[width=1.0\textwidth]{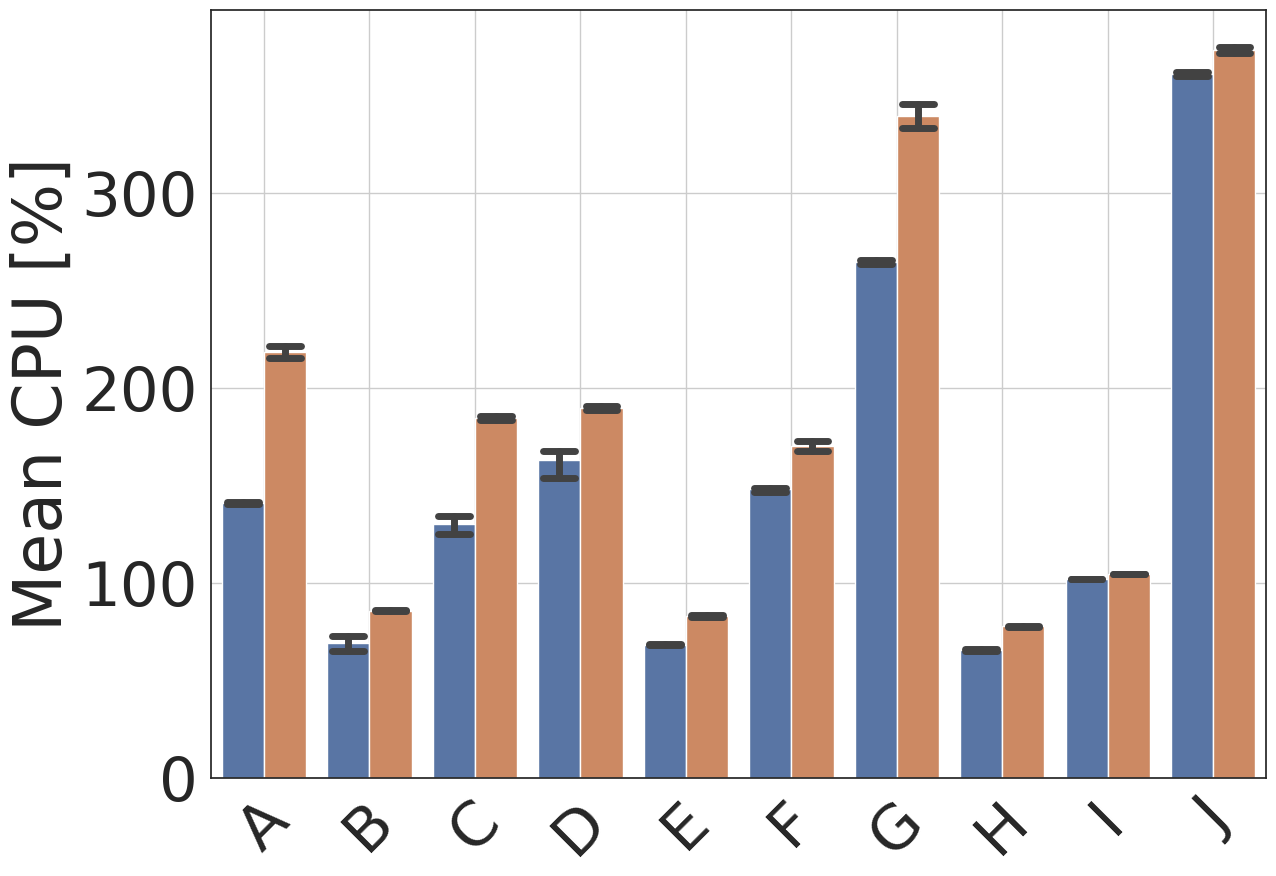}
      \label{fig:exp:mean_cpu}
    \end{subfigure}
        \begin{subfigure}{0.32\textwidth}
      \centering
      \includegraphics[width=1.0\textwidth]{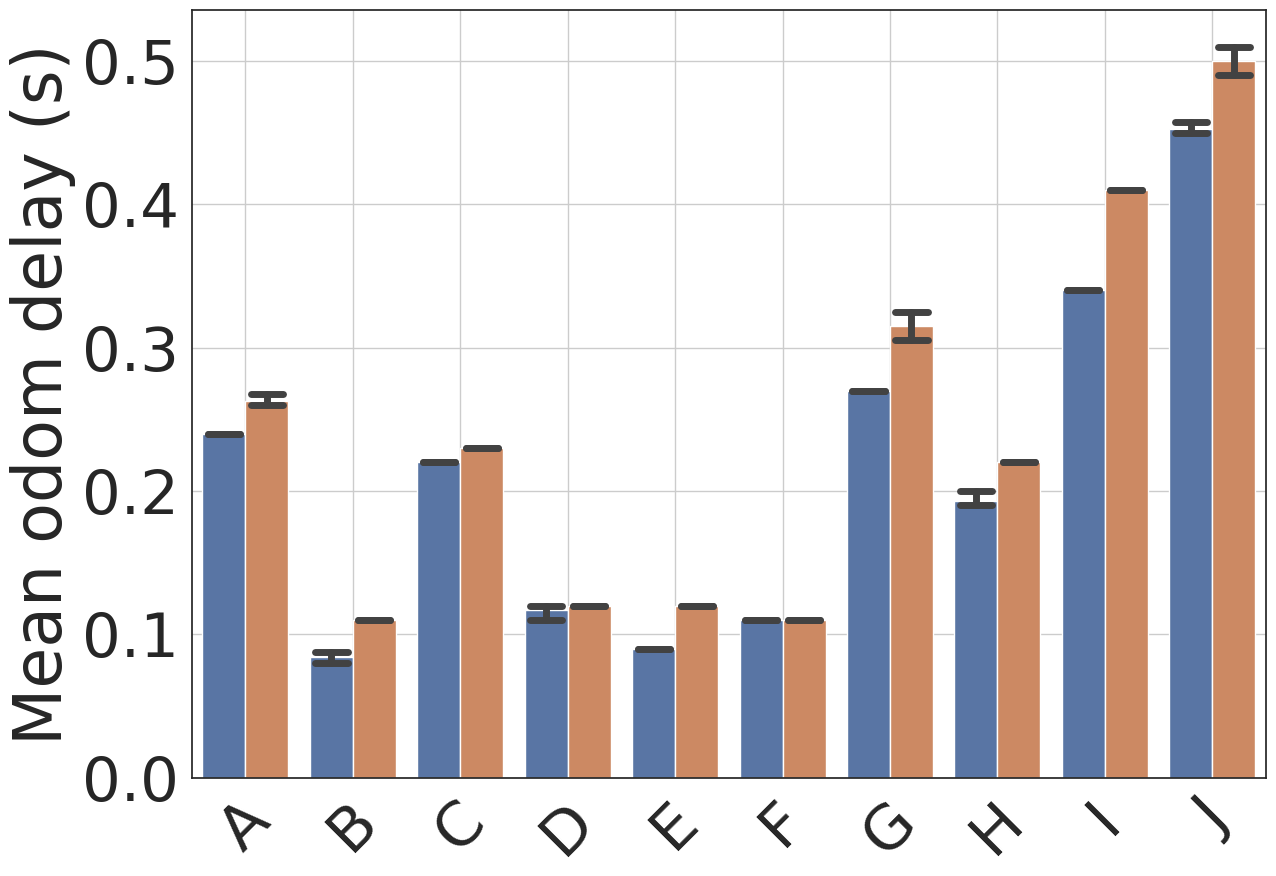}
      \label{fig:exp:max_cpu}
    \end{subfigure}
    \begin{subfigure}{0.32\textwidth}
      \centering
      \includegraphics[width=1.0\textwidth]{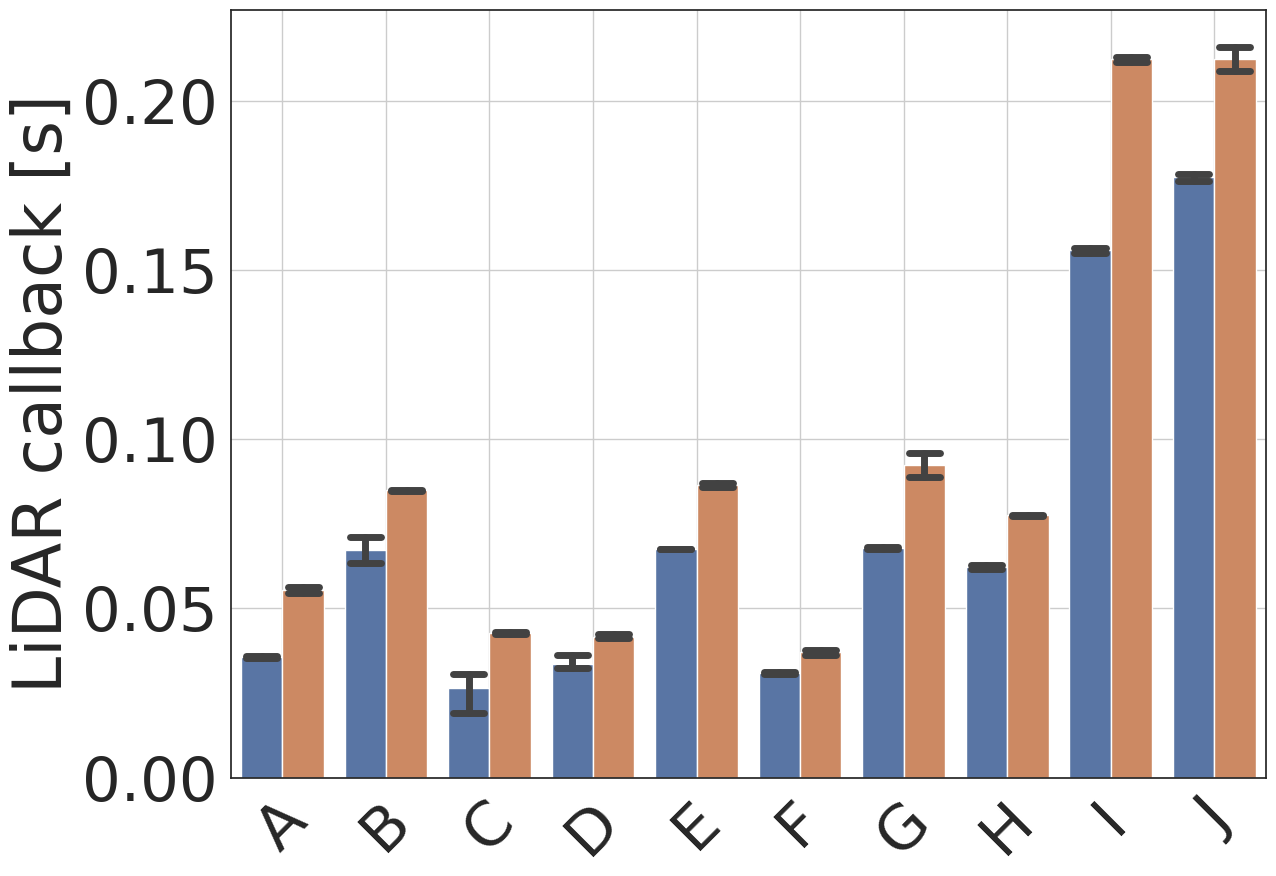}
      \label{fig:exp:mean_odom}
    \end{subfigure}
        \begin{subfigure}{0.32\textwidth}
      \centering
      \includegraphics[width=1.0\textwidth]{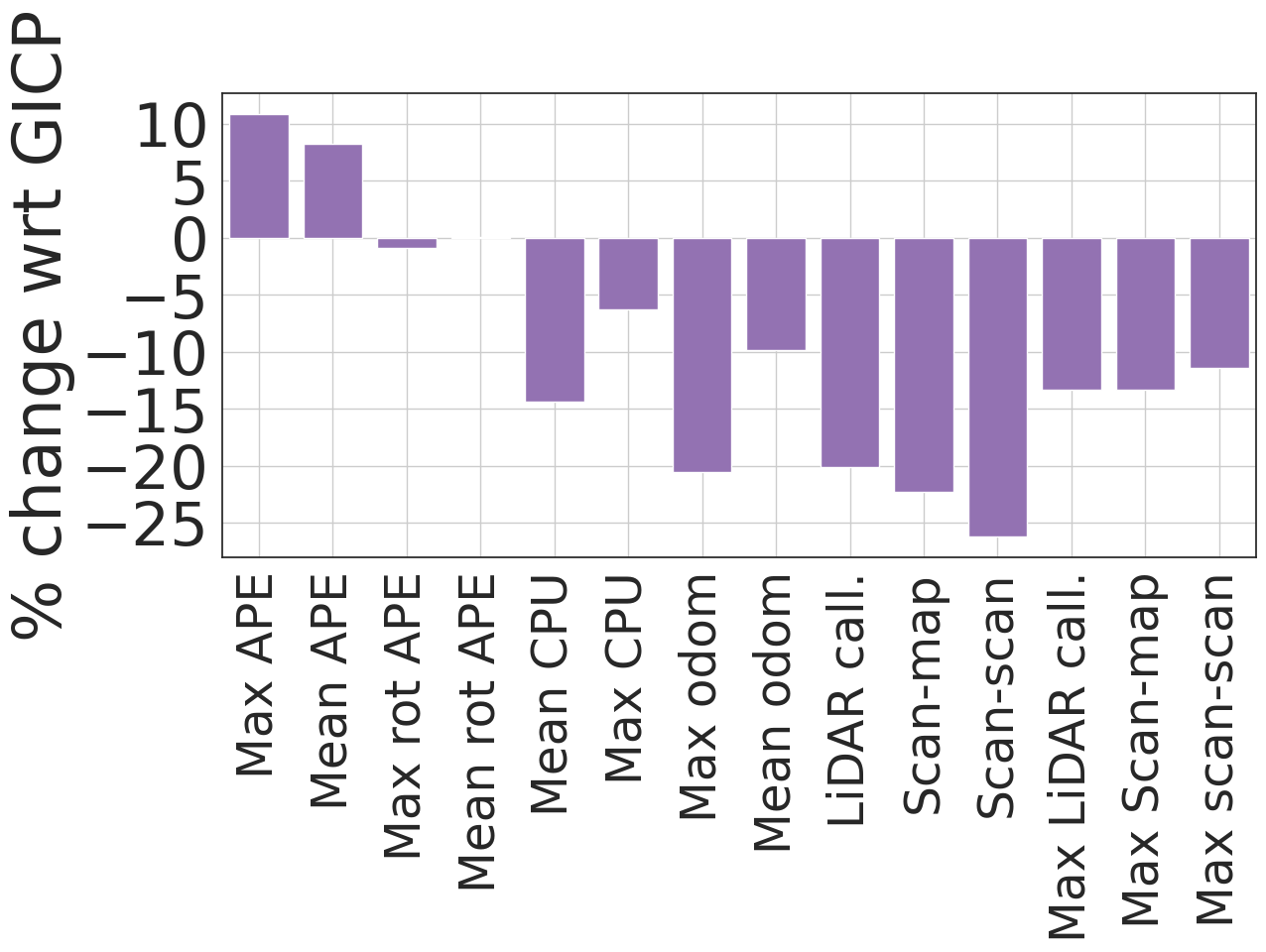}
      \label{fig:ex:lidar_callback}
    \end{subfigure}
\vspace{-0.5cm}
    \caption{\footnotesize{Results of \textit{GICP from normals} and GICP comparison in LOCUS 2.0. \matteo{For the meaning of the labels A-J  see \tabref{tab:tabledataset}}}}
    \label{fig:comparison_gicp}
\end{figure*}

The \figref{fig:comparison_gicp}.a-e present the comparison results between \textit{GICP from normals} and GICP across datasets, while \figref{fig:comparison_gicp}.f shows the average percentage change across all dataset for each metric with respect to the GICP method.
\textit{GICP from normals} reduces all the computational metrics in LOCUS 2.0: mean and max CPU usage, mean and max odometry delay, scan-to-scan, scan-to-submap, \lidar callback duration and their maximum times.
The computation burden is, on average, reduced by $18.57\%$ for all those metrics and datasets. 
This reduction benefits the odometry update rate since frequency increases by $11.10\%$ that is beneficial for another part of the system, i.e. for path planning and control algorithms. \morrell{The \lidar callback is generally higher for datasets I and J largely due to the consistently large cross-section of Kentucky Underground making GICP take longer.}
One drawback of this method leads to slight increase in the mean and max APE errors. 
The reason is that normals are calculated from sparse point clouds, and those normals are stored in the map without any further recomputation. In GICP, the covariances are recalculated from dense map point clouds.
The mean and max APE error increases across all datasets on average $10.82\%$, but the increase is not for all datasets.
Without including the tunnel dataset (\textbf{F}) the average APE error is only $5.23\%$.
The rotational APE errors do not change much since max APE decreases $0.94\%$, while mean APE increases $0.1\%$.

\subsubsection{Adaptive Voxel Grid Filter}
\label{sec:exp:adaptive_voxelization}
\begin{figure}
	\centering
	\includegraphics[width=1.0\columnwidth]{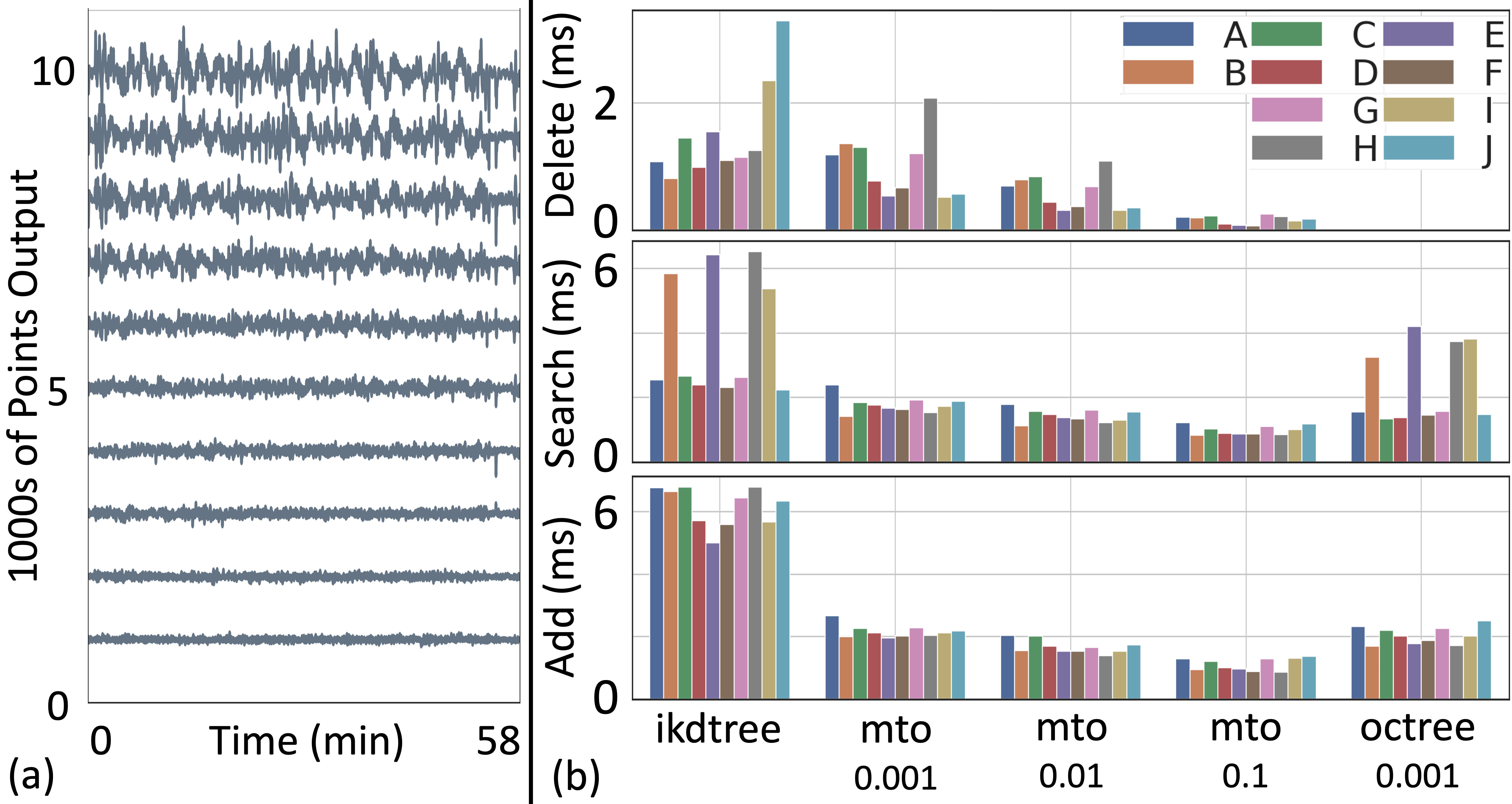}
    \caption{\footnotesize{\morrell{(a) Number of points after the adaptive voxel grid filter for different set-points (on dataset I). (b) Timeplots for deleting, adding, searching for different map storage mechanisms.}}}
    \label{fig:exp:delete_search_add_duration}
    \label{fig:keeping_points_adaptive_voxelization}
\end{figure}

The second experiment presented in this section shows LOCUS 2.0 adaptive behavior.
The experiments are run across all datasets with \textit{GICP from normals} and the same parameters as in \secref{sec:gicp_from_normals} with an \ikdtree~data structure for map maintenance with a box size $50~m$, and $N_{desired}$ ranging from $1000$ to $10000$.
\matteo{\figref{fig:keeping_points_adaptive_voxelization}.a shows how the adaptive voxel grid filter keeps the number of points relatively consistent across a 1~hour dataset, no matter what the input $N_{desired}$ is.}

\begin{figure}
    \centering
    \begin{subfigure}{0.24\textwidth}
      \centering
      \includegraphics[width=1.0\textwidth]{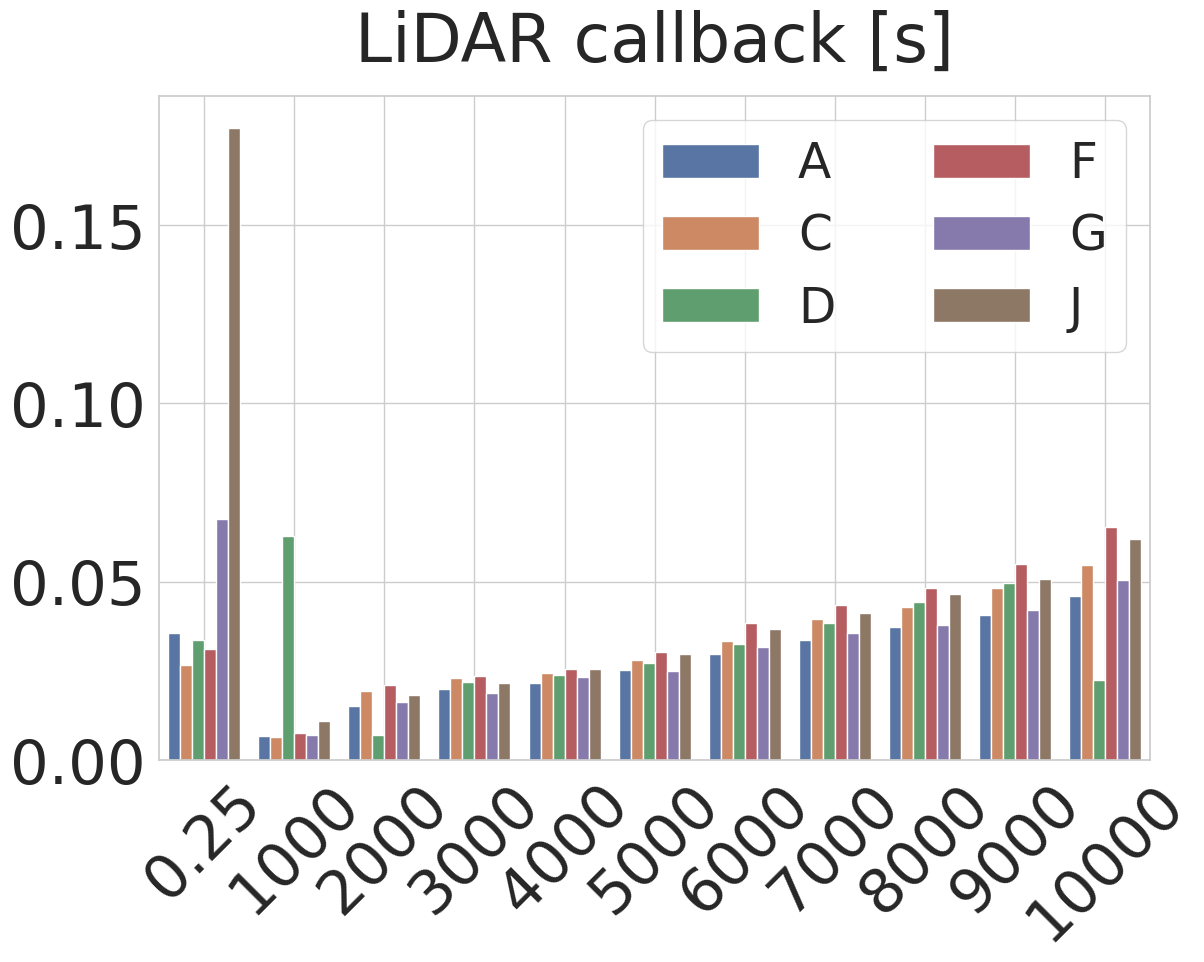}
    \end{subfigure}\hfill
    \begin{subfigure}{0.24\textwidth}
      \centering
      \includegraphics[width=1.0\textwidth]{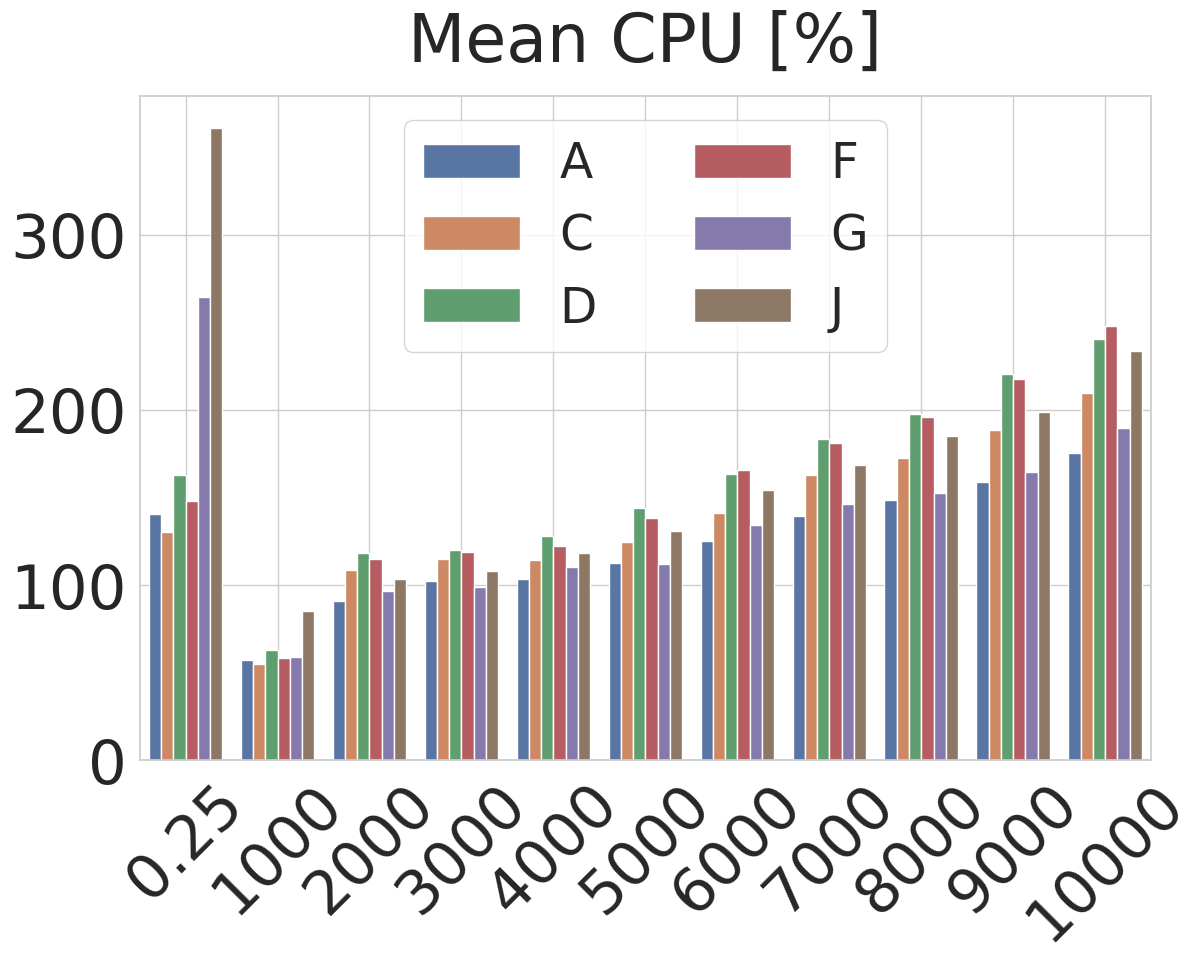}
    \end{subfigure}
    \caption{\footnotesize{The figure presents comparison between adaptive voxelization $1000-10000$ points and constant leaf size $0.25$  \matteo{(our previously used static voxel size). For $0.25$ voxel the average number points for each dataset (A, C, D, F, G, J) is $8423$, $5658$, $2967$, $2368$, $8511$, $15901$. }}
    }
    \label{fig:exp:adaptive_voxelization_variability}
\end{figure}

\begin{figure}
	\hspace{-0.5cm}
    \begin{subfigure}{0.25\textwidth}
      \includegraphics[width=1.0\textwidth]{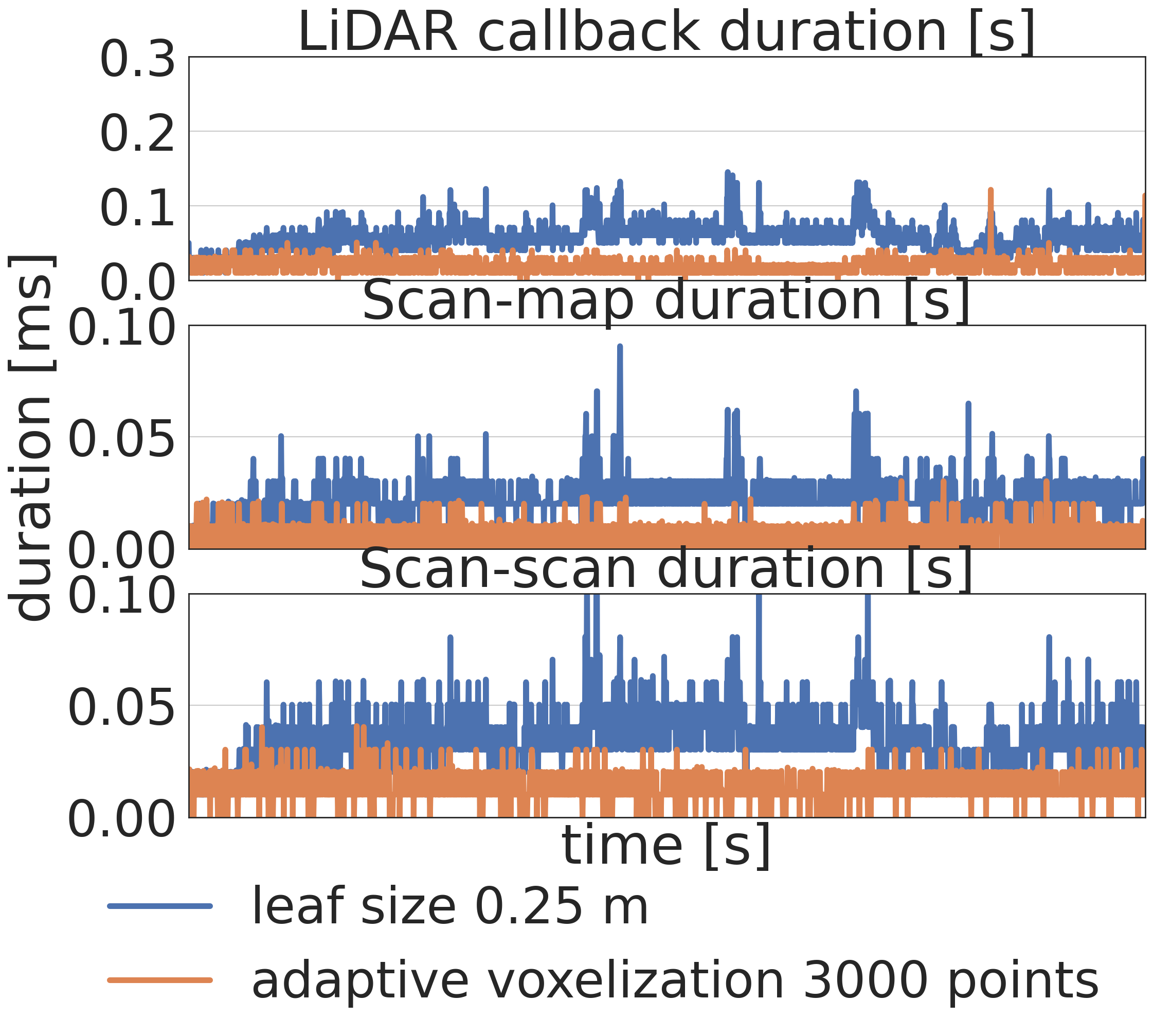}
      \caption{\footnotesize{Husky urban dataset (A).}}
    \end{subfigure} 
    \begin{subfigure}{0.25\textwidth}
      \includegraphics[width=1.0\textwidth]{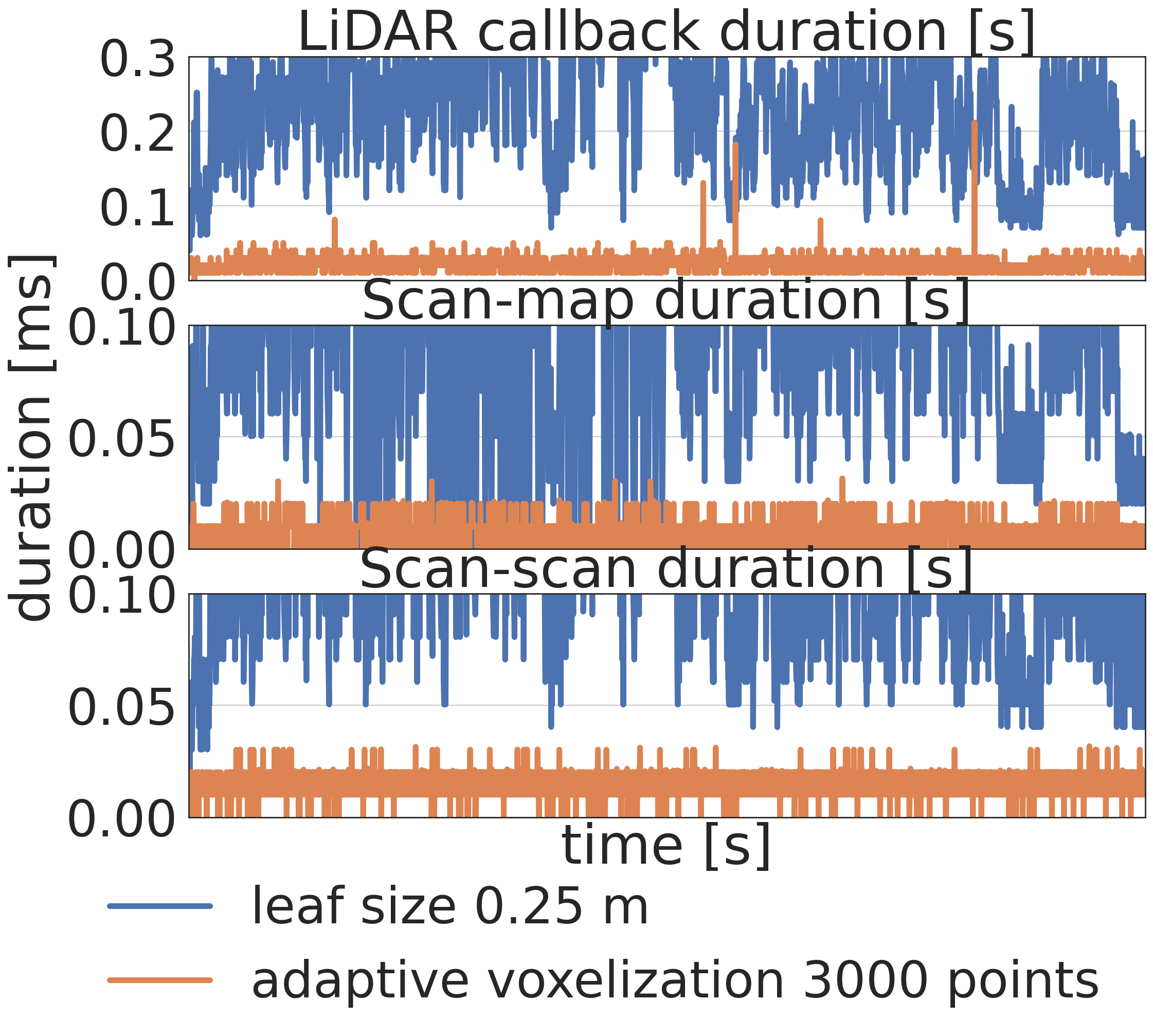}
      \caption{\footnotesize{Husky cave dataset (J).}}
    \end{subfigure}
    \caption{\footnotesize{\morrell{A comparison of the consistency of computation time for adaptive voxelization with $3000$ points and constant leaf size $0.25$ (our previously used static voxel size). a) Urban dataset using $2$ \lidars. b) Cave dataset using $3$ \lidars. }}}
    \label{fig:exp:adaptive_plots_3000}
\end{figure}

\morrell{There is still some variability in the computation time across a dataset, though. Nonetheless, as shown in \figref{fig:exp:adaptive_plots_3000}, the approach produces a consistent average computation time across different environments and sensor configurations, without any large spikes in computation time. This performance gives more predictable computational loads, regardless of robot or environment, as further reinforced in~\figref{fig:exp:adaptive_voxelization_variability}, where the average callback time and CPU load are similar for all datasets at the same adaptive voxelization setting.
}

\subsection{Memory}
\subsubsection{Map maintenance}

The third experiment presented in this section presents the benefit of sliding-window maps in real-time systems compared to classic, static octree-based structures.
In these experiments, LOCUS 2.0 uses: \ikdtree and multi-threaded octree (mto).
The octree with leaf size $0.001~m$ is the baseline used in LOCUS 1.0 to maintain full map information. 
To assess different parameters mto runs with leaf size $0.1~m$, $0.01~m$, and $0.001~m$. 

\begin{figure}[t]
\vspace{0.5em}
	\centering
	\begin{subfigure}{0.5\textwidth}
		\centering
		\includegraphics[width=0.485\textwidth]{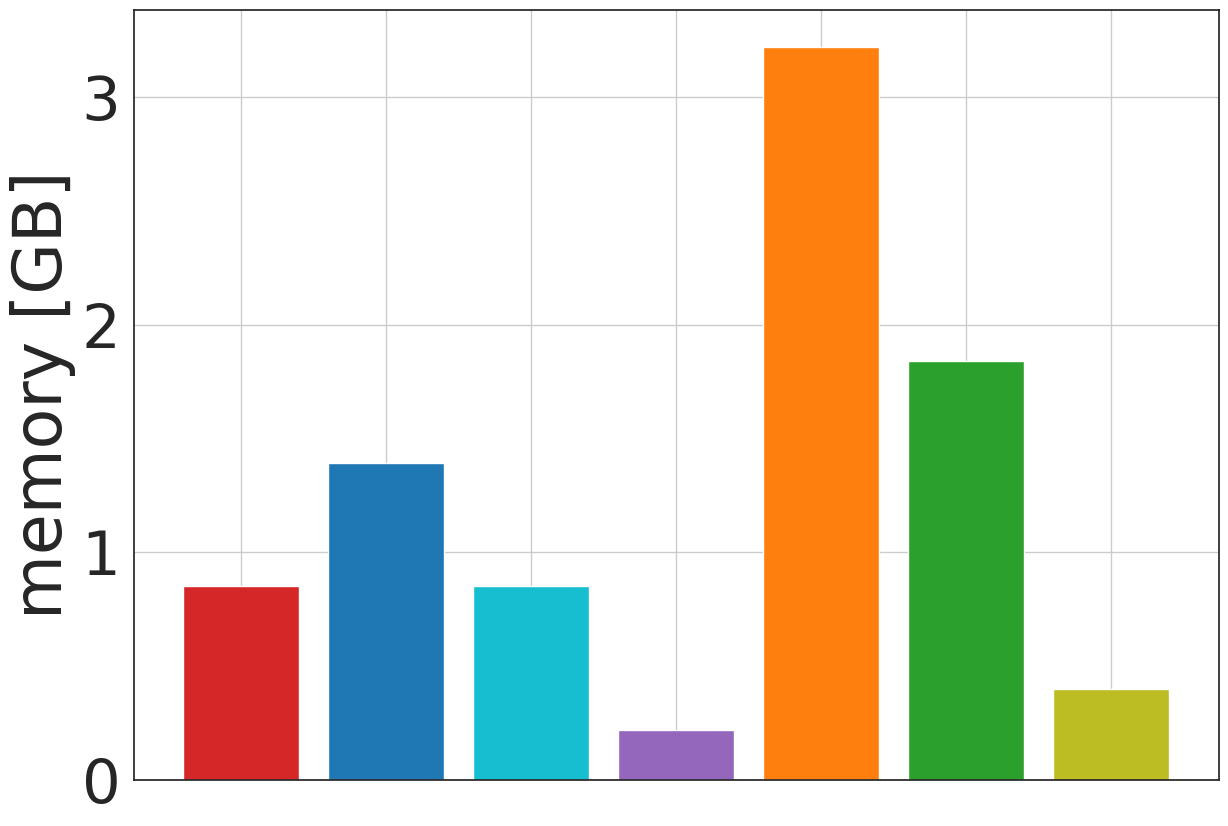}%
		\hfill
		\includegraphics[width=0.51\textwidth]{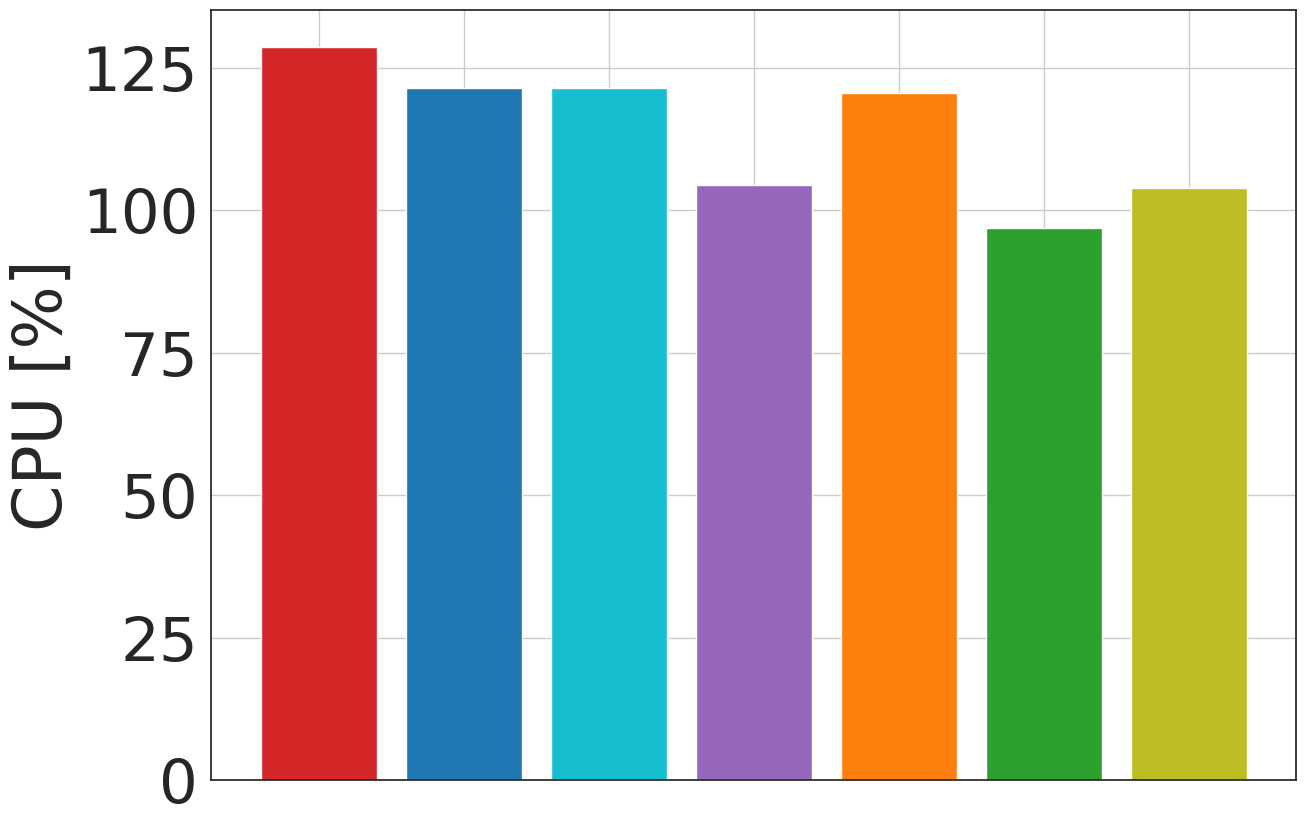}
		\caption{\footnotesize{Husky tunnel (F).}}
	\end{subfigure}
	\begin{subfigure}{0.5\textwidth}
	\centering
	\includegraphics[width=0.5\textwidth]{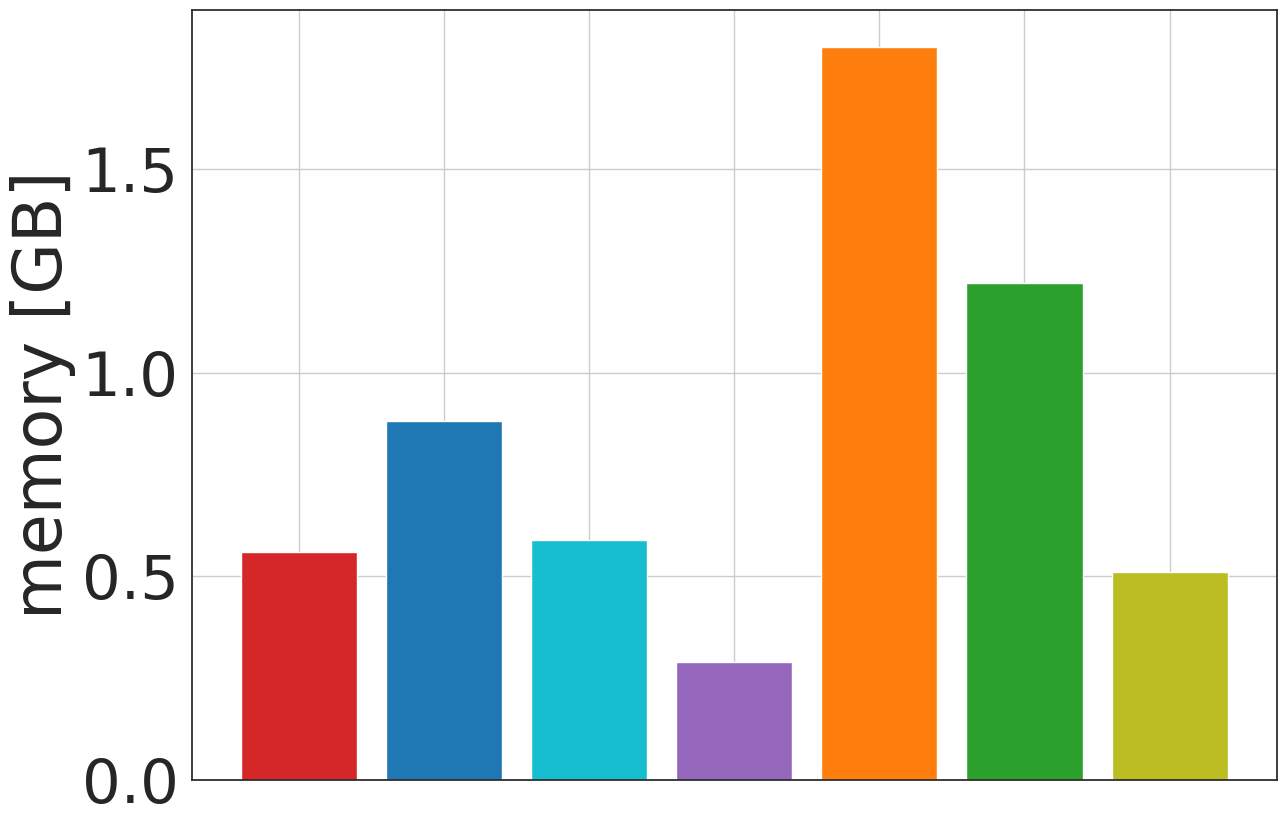}%
	\hfill
	\includegraphics[width=0.5\textwidth]{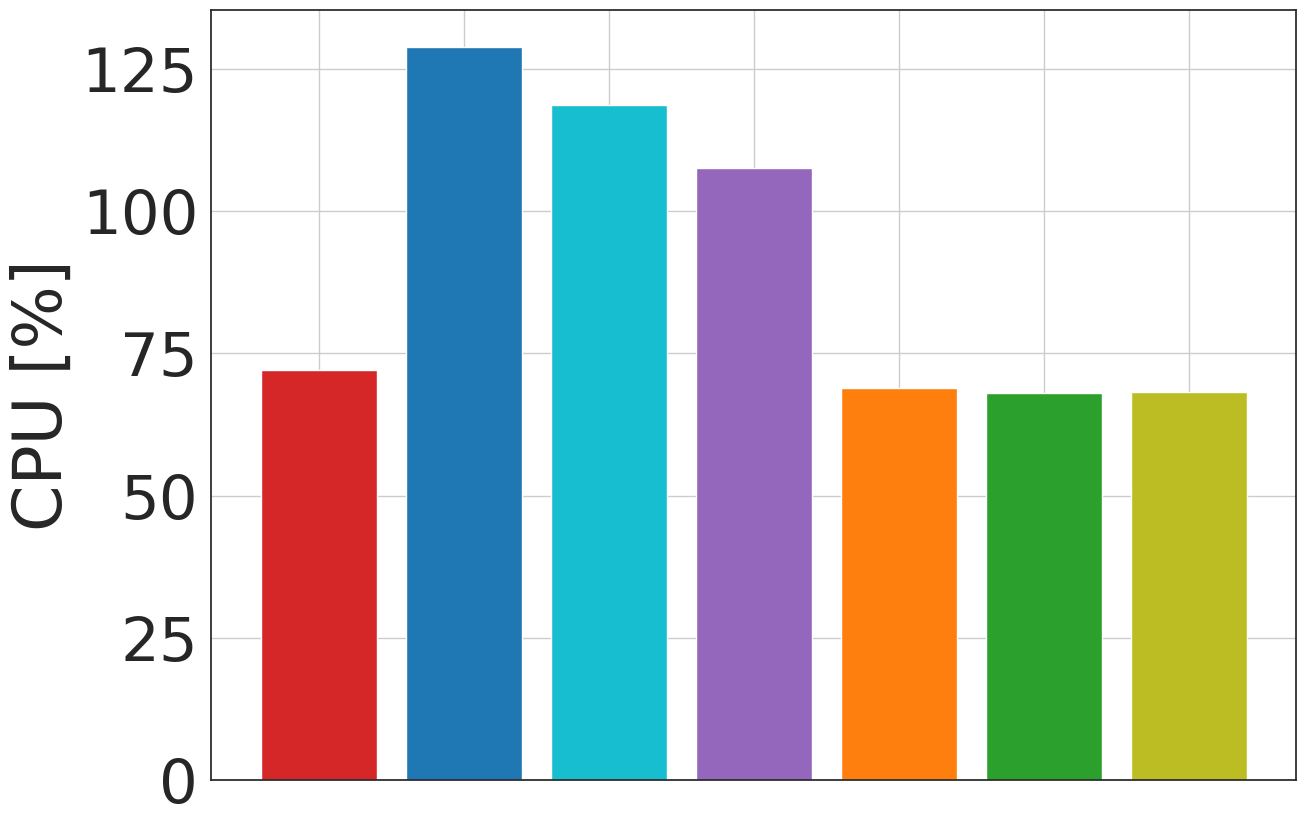}
	 \vspace{-0.6cm}
	\caption{\footnotesize{Spot cave (I).}}
\end{subfigure}
\begin{subfigure}{0.5\textwidth}
    \begin{subfigure}{0.49\textwidth}
	\centering
	\includegraphics[width=1\textwidth]{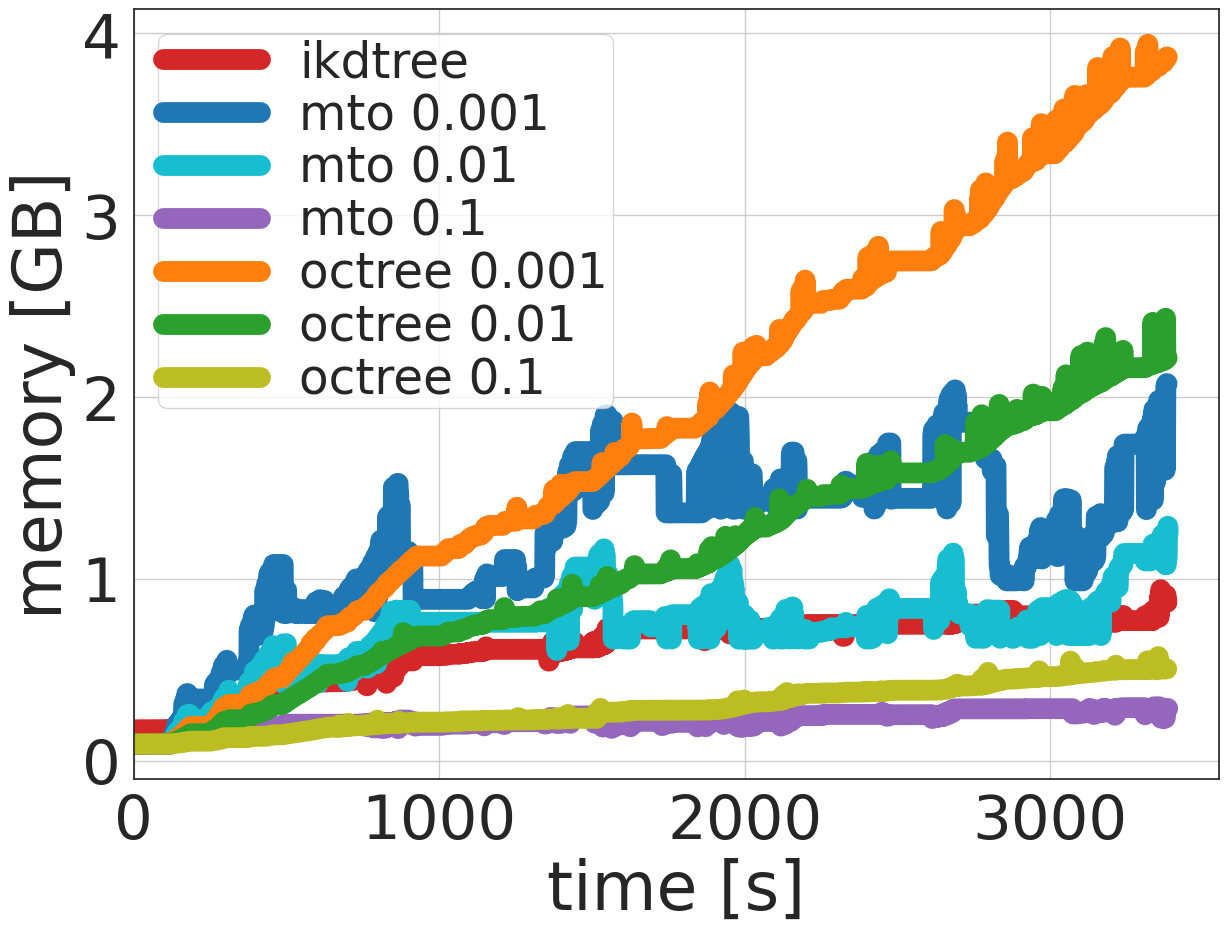}
	\vspace{-0.5cm}
	\caption{\footnotesize{Memory in time for (F).}}
	\label{fig:map_in_time_husky}
\end{subfigure}
\hfill
\begin{subfigure}{0.49\textwidth}
	\centering
	\includegraphics[width=1\textwidth]{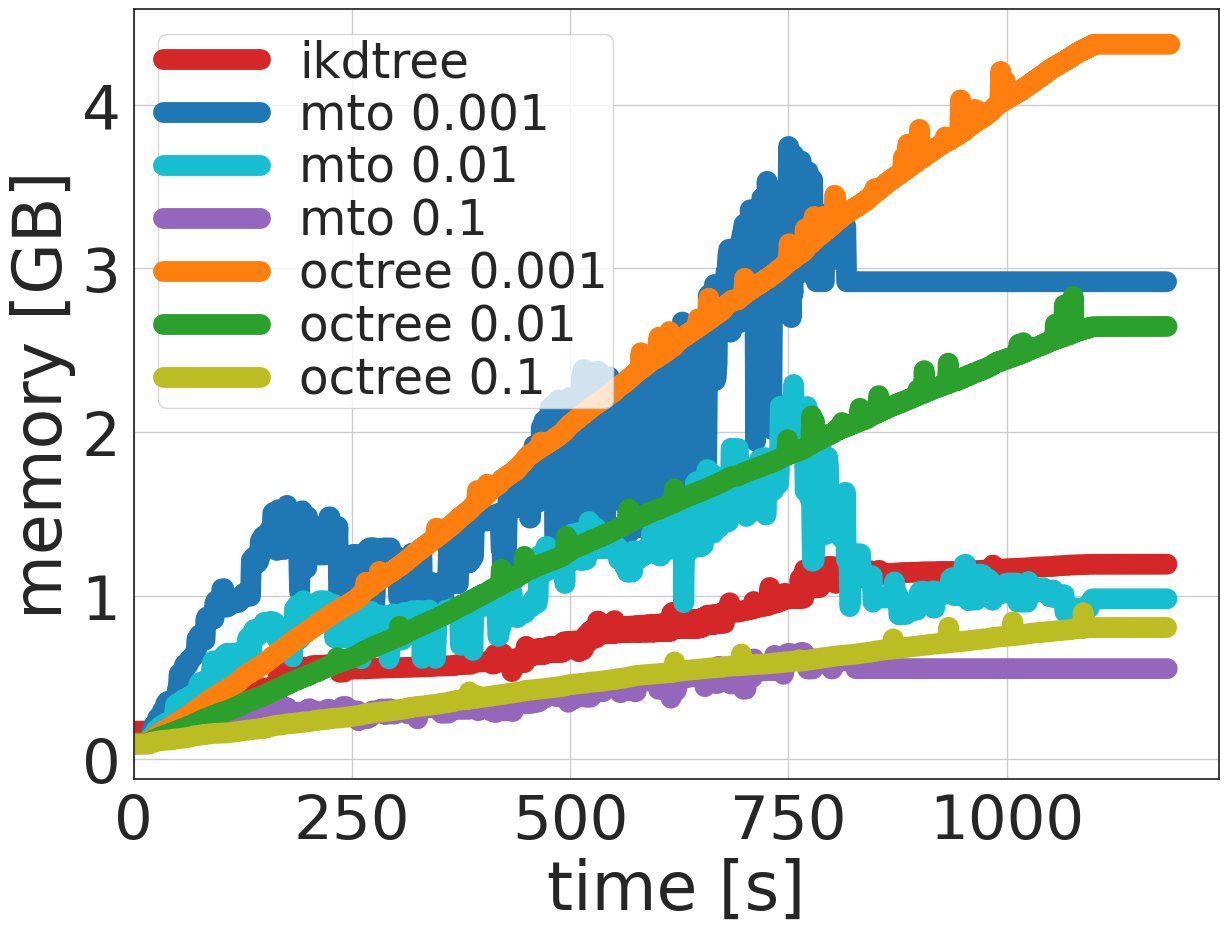}
	\vspace{-0.5cm}
	\caption{\footnotesize{Memory in time for (I).}}
	\label{fig:map_in_time_spot}
\end{subfigure}
\end{subfigure}
\vspace{-0.2cm}
	\caption{\footnotesize{The plots show the metrics for different datasets relative to the number of fixed voxelized points.}}
	\label{fig:exp:mapping_techniques}
\end{figure}

For sliding-window approaches, the map size is $50~m$ since it is the maximum range of \lidars.
For scan-to-scan and scan-to-submap stage \textit{GICP from normals} is used with the parameters chosen based on previous experiments.
\figref{fig:exp:mapping_techniques} presents the maximum memory use for \textbf{F} and \textbf{I} dataset and how memory occupation evolves over time.
The largest memory occupancy is for octree and mto version for  $0.001~m$ leaf size. The \ikdtree~achieves similar performance in terms of memory and CPU usage as the mto  with leaf size $0.01m$. 
\tabref{tab:map_comparison} shows how sliding-window map approaches reduce the memory usage while increasing the CPU usage in comparison to the reference method from LOCUS 1.0.

\begin{table}[t]
\vspace{0.5em}
\caption{\footnotesize{Relative memory and CPU change.
	}}
	\label{tab:map_comparison}
	\begin{center}
		\resizebox{1.0\linewidth}{!}{
			\begin{tabular}{c c c c c c } \hline
				 & \ikdtree & 	mto $0.001$ & mto $0.01$ & mto $0.1$ & octree $0.001$ \\ 
				\toprule
				Memory  & $-68.09\%$ & $-38.88\%$ & $-62.15\%$ & $-87.76\%$  & X \\ 
				\midrule
				CPU & $9.36\%$ & $50.42\%$ & $44.36\%$ & $19.61\%$ & X \\ 
				\bottomrule
			\end{tabular}
		}
	\end{center}
\end{table}

\matteo{\figref{fig:exp:delete_search_add_duration}.b shows the deletion, insertion and searching procedure timings for different mapping strategies. }
\BIGIkdtree has the most time-consuming procedures for searching, deletion, and insertion across all datasets.
For insertion, \ikdtree~uses on average $222\%$ more computation time than the octree data structure as new points need to be stored in a particular manner.
For search, \ikdtree~gives on average $140\%$ more computation than the octree data structure. 

\subsubsection{Map size}

\begin{figure}[]
\vspace{0.5em}
\centering
\begin{subfigure}{0.23\textwidth}
    \centering
    \includegraphics[width=1.0\textwidth]{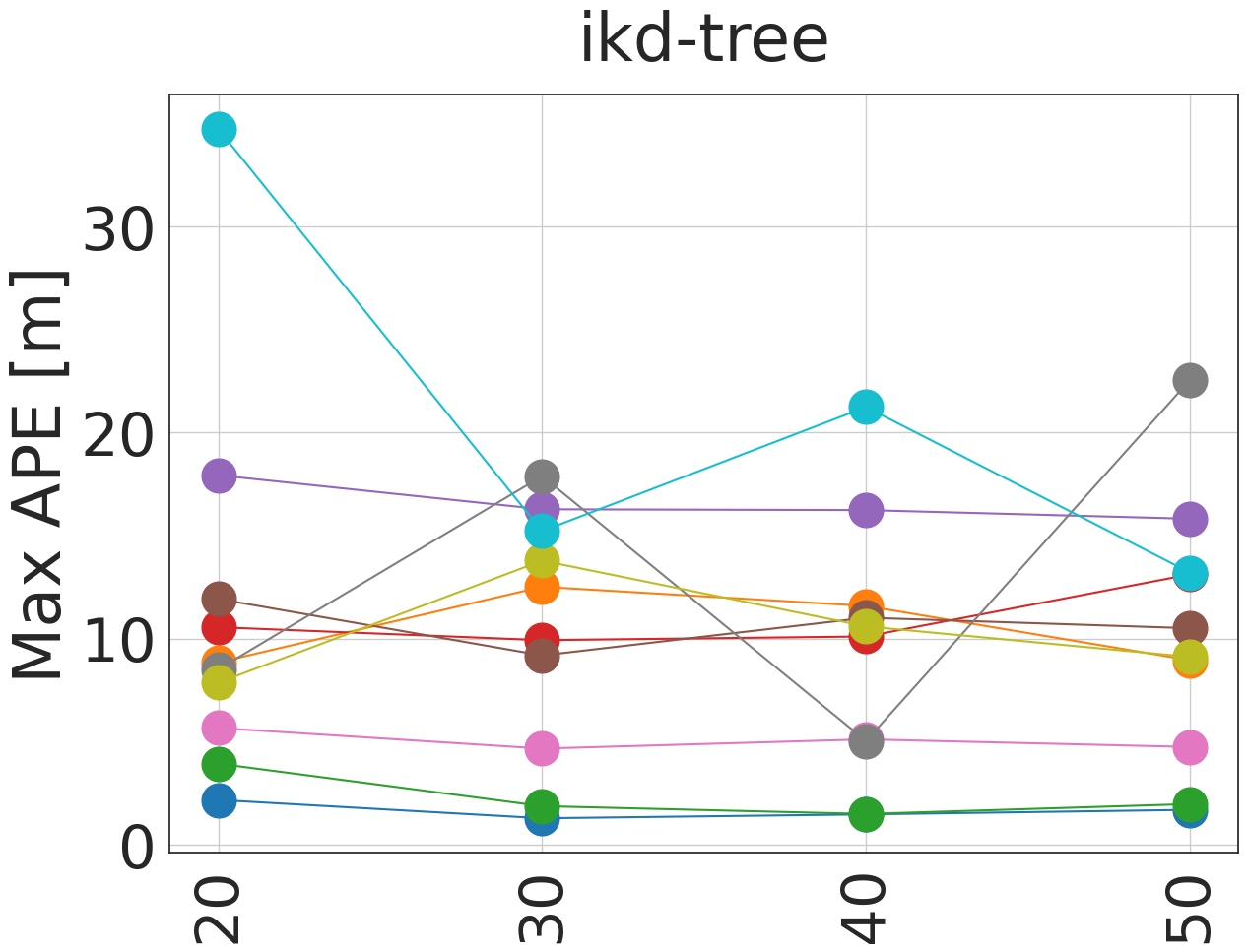}
 \end{subfigure}
\hfill
\begin{subfigure}{0.23\textwidth}
    \centering 
    \includegraphics[width=1.0\textwidth]{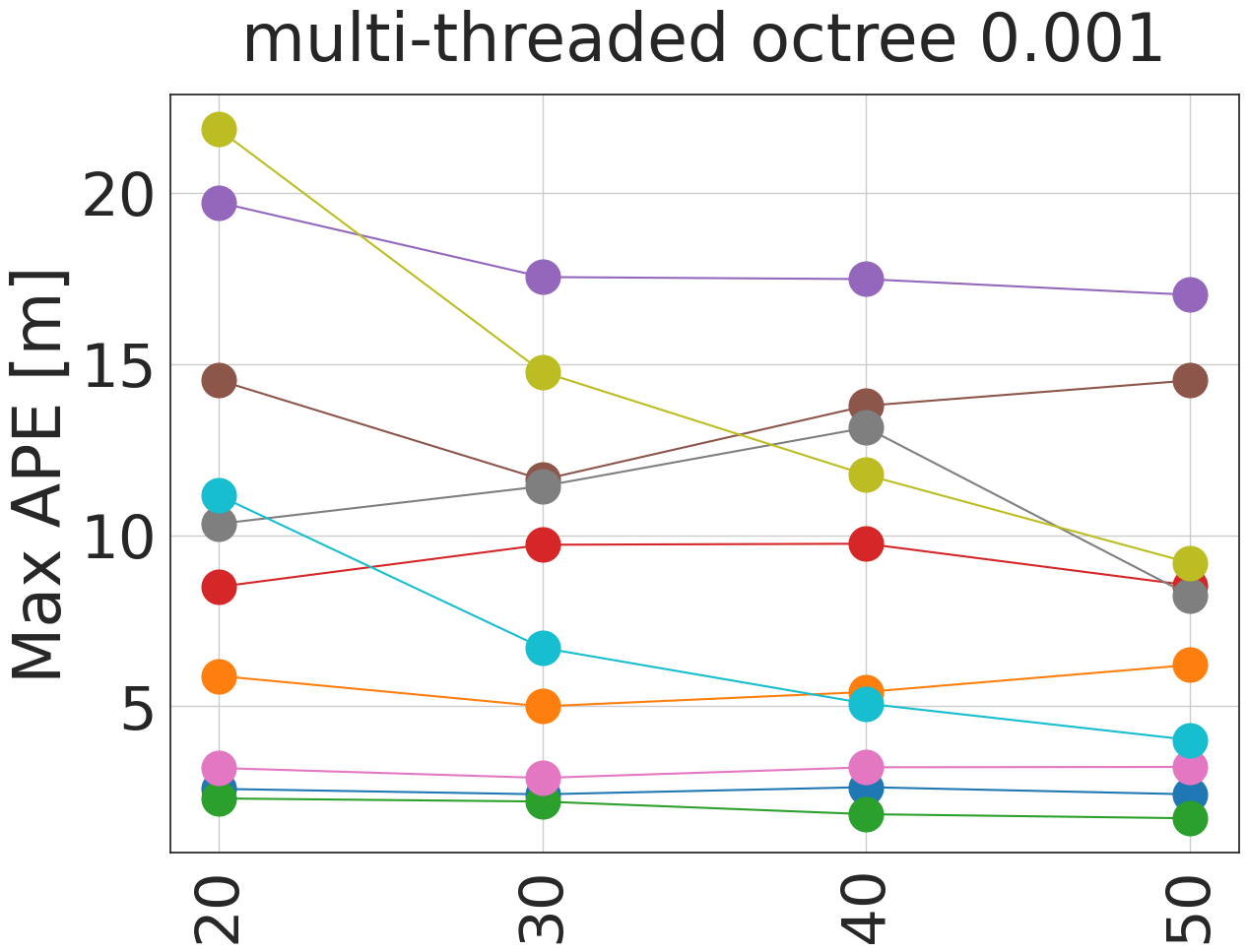}
 \end{subfigure}
\begin{subfigure}{0.23\textwidth}
    \centering
    \includegraphics[width=1.0\textwidth]{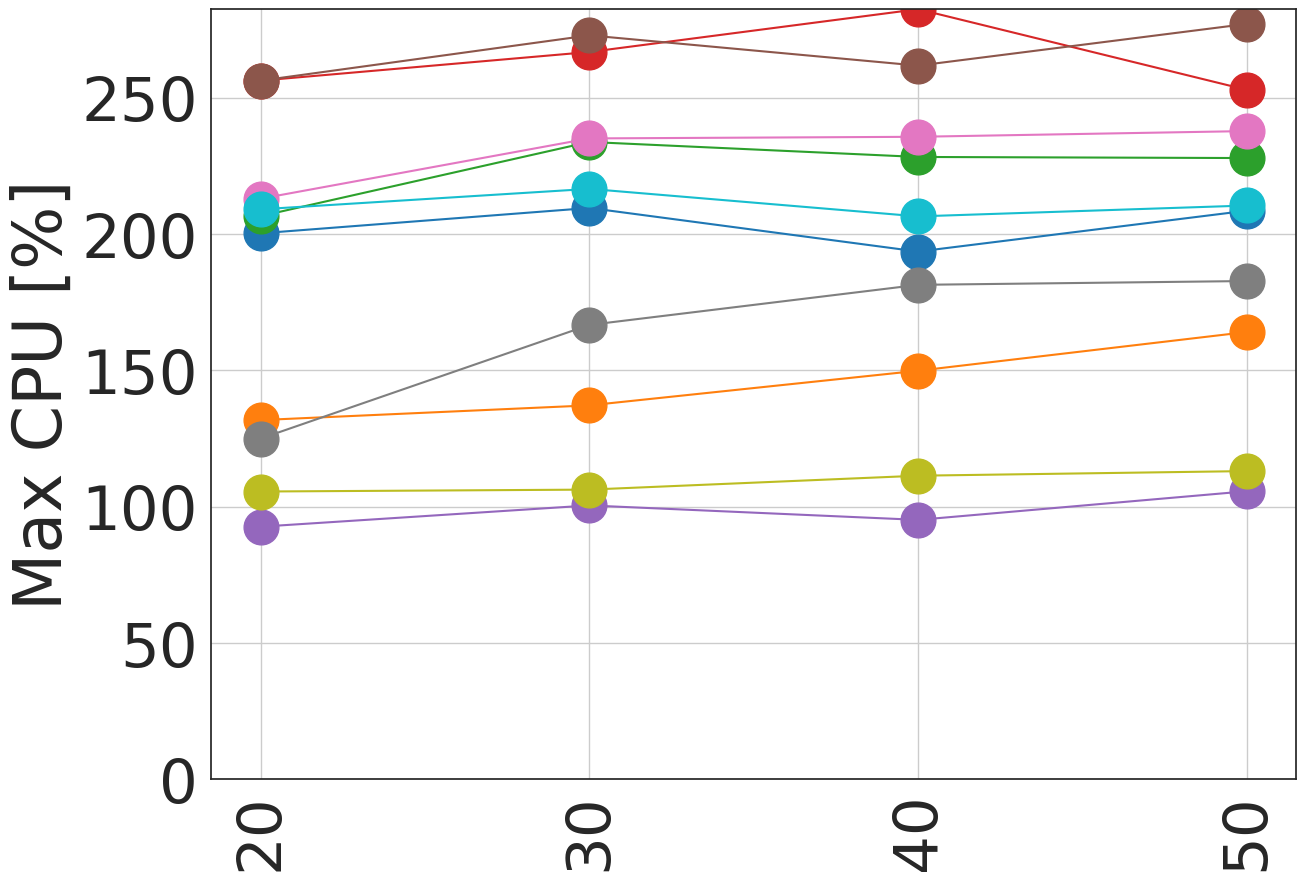}
\end{subfigure}
\hfill
\begin{subfigure}{0.23\textwidth}
    \centering
    \includegraphics[width=1.0\textwidth]{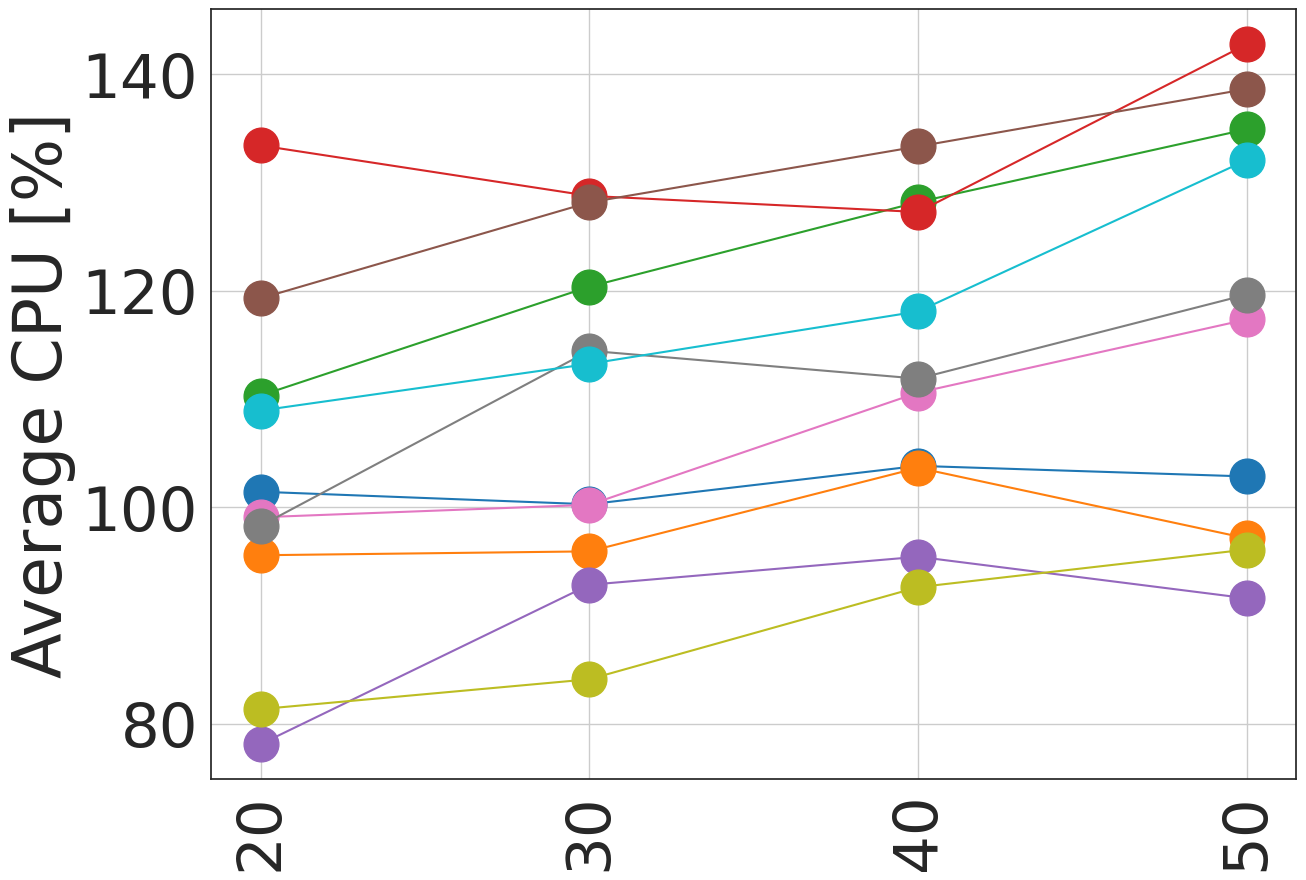}
\end{subfigure}
\begin{subfigure}{0.23\textwidth}
    \centering
    \includegraphics[width=1.0\textwidth]{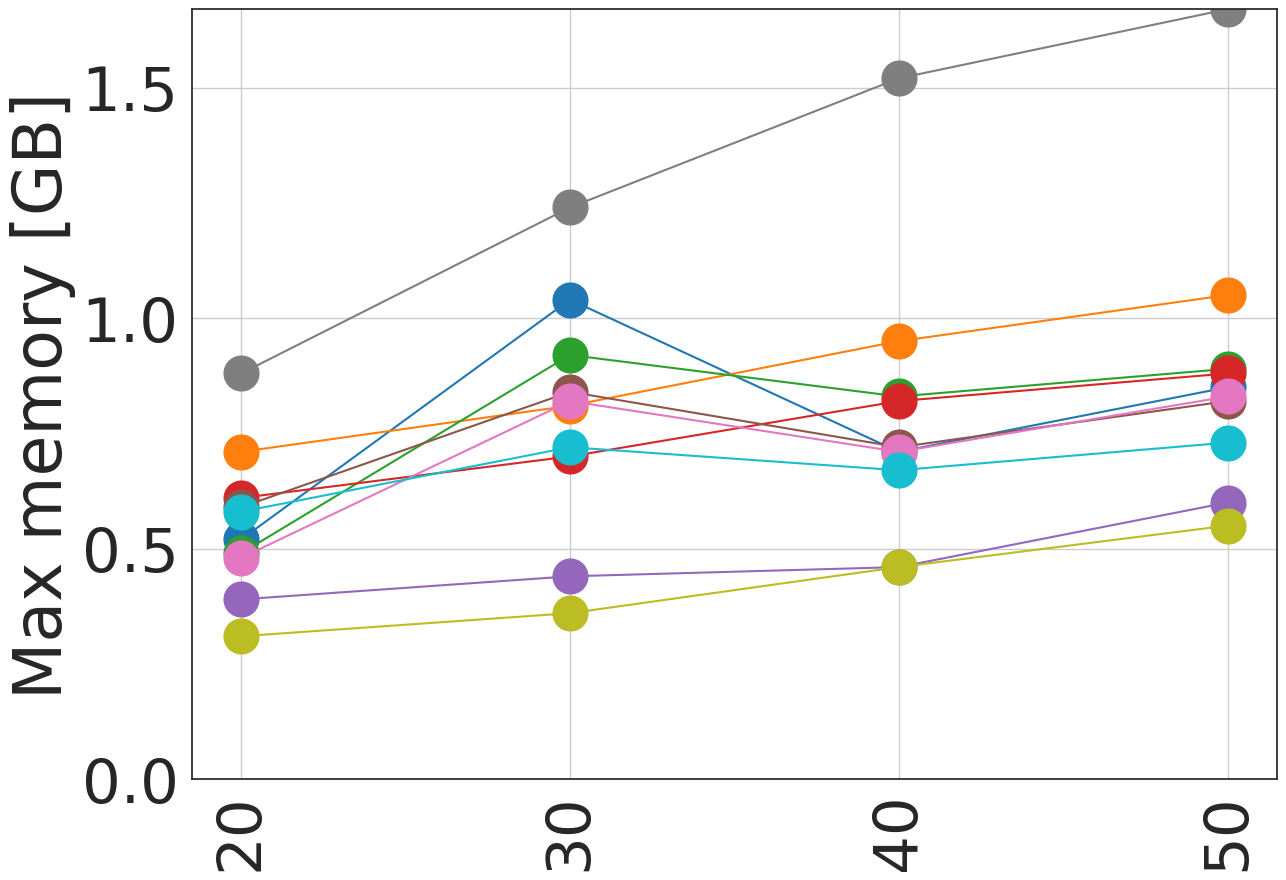}
 \end{subfigure}
\hfill
\begin{subfigure}{0.217\textwidth}
    \centering
    \includegraphics[width=1.0\textwidth]{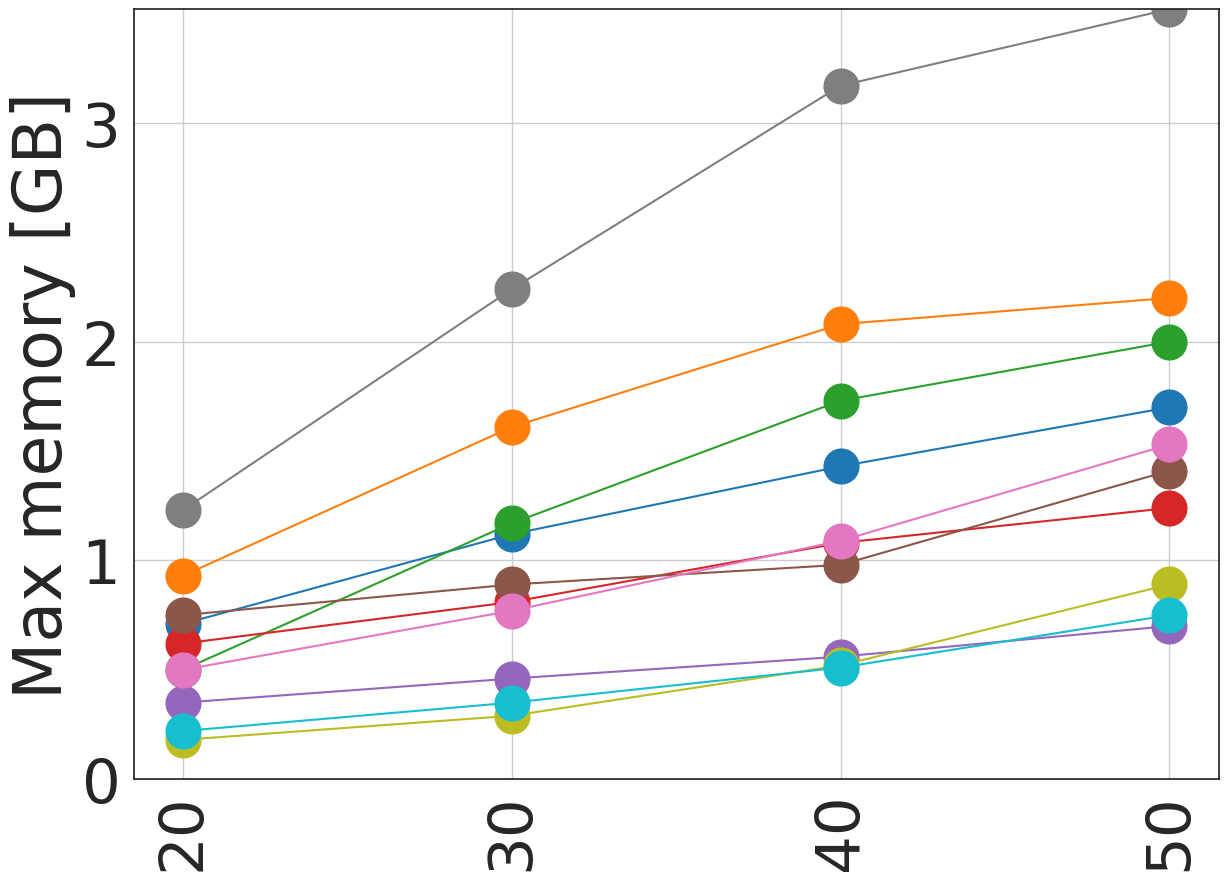}
 \end{subfigure}
\begin{subfigure}{0.5\textwidth}
	\centering
	\includegraphics[width=1.0\textwidth]{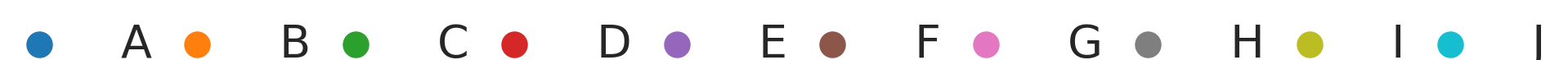}
\end{subfigure}
\caption{\footnotesize{Results based on size of the map for \ikdtree~and mto $0.001$ across all datasets (A-J).}}
\label{fig:exp:map_size}
\end{figure}

These experiments show that the size of the sliding map is an important parameter to consider while trading off the computational, memory load and accuracy of the result.
The sliding-window map allows the system to be bounded by the maximal memory that the robot allocates, following the paradigms of Real-Time Operating System requirements.
\figref{fig:exp:map_size} shows the max APE, CPU and memory metrics for \ikdtree~and mto in terms of map size.
The smaller map size gives the robot a lower upper bound for the memory, but on the other hand, instances with larger maps have lower APE as there is more overlap between scan and map. Other than memory, these larger maps also see larger the mean and max CPU load.

\matteo{
\subsection{Comparison to the state-of-the-art}

\matteo{
\tabref{tab:state_of_the_art_comparison} shows the comparison study for LOCUS 2.0 against the state-of-the-art methods FAST-LIO \cite{xu2021fast} and LINS \cite{qin2020lins} for different environment domains: urban, tunnel, cave (A,C,F,H,I,J).
The table shows that LOCUS 2.0 presents a state-of-the-art performance in terms of max and mean APE error metrics and achieves the smallest errors in $5$ out of $6$ presented datasets. 
In addition, LOCUS 2.0 is the only method that does not fail in the tunnel type of the environment (dataset F) where \lidar slip occurs. In terms of computation, LOCUS 2.0 achieves equivalent performance to FAST-LIO. Memory usage for LOCUS 2.0 is slightly larger, yet this is likely related to the resolution of the map chosen by default in all the systems.
}
\vspace{-1em}
\begin{table}[t]
\caption{\footnotesize{\matteo{Comparison of LOCUS 2.0 to the state-of-the art methods FAST-LIO
\vspace{-1em}
\cite{xu2021fast} and LINS \cite{qin2020lins} for different environment domains.}
	}}
	\label{tab:state_of_the_art_comparison}
	\begin{center}
		\resizebox{1\linewidth}{!}{
			\begin{tabular}{c  c | c c | c c |  c } \hline
Dataset & Algorithms &  \multicolumn{2}{c}{APE } &  \multicolumn{2}{c}{CPU [\%]} & max memory \\ \toprule
 &  & max [m] & mean [\%]  & max & mean & [GB] \\ 
 \midrule
& LOCUS 2.0  & $\textbf{0.19}$ & $\textbf{0.09}$ & $102.38$ & $185.50$ & $1.06$\\ 
A & FAST-LIO & $0.79$ & $0.30$ & $89.11$ & $126.40$ & $\textbf{0.36}$\\
& LINS & $0.43$ & $0.18$ & $\textbf{40.84}$ & $\textbf{81.50}$ & $0.42$ \\
\bottomrule
& LOCUS 2.0  & $\textbf{0.16}$ & $\textbf{0.24}$ & $114.79$ & $198.00$ & $1.30$\\ 
C & FAST-LIO & $2.21$ & $4.22$ & $76.46$ & $307.20$ & $0.99$\\
& LINS & $0.43$ & $0.60$ & $\textbf{38.43}$ & $\textbf{75.30}$ & $\textbf{0.47}$ \\
\bottomrule
& LOCUS 2.0  & $\textbf{0.67}$ & $\textbf{0.45}$ & $119.00$ & $229.20$ & $1.98$\\ 
F & FAST-LIO & $48555.33$ & $9268.71$ & $156.73$ & $401.30$ & $11.31$\\
& LINS & $52.73$ & $23.35$ & $\textbf{28.10}$ & $\textbf{52.30}$ & $\textbf{0.47}$ \\
\bottomrule
& LOCUS 2.0  & $\textbf{0.57}$ & $\textbf{0.23}$ & $61.05$ & $169.90$ & $2.42$\\ 
H & FAST-LIO & $5.92$ & $5.69$ & $75.15$ & $160.80$ & $0.62$\\
& LINS & $12.11$ & $8.05$ & $\textbf{39.19}$ & $\textbf{97.90}$ & $\textbf{0.61}$ \\
\bottomrule
& LOCUS 2.0  & $1.39$ & $1.95$ & $\textbf{72.11}$ & $141.60$ & $1.01$\\ 
I & FAST-LIO & $0.99$ & $1.44$ & $117.87$ & $167.80$ & $\textbf{0.80}$\\
& LINS & $\textbf{0.86}$ & $\textbf{0.85}$ & $75.90 $ & $\textbf{101.40}$ & $0.85$ \\
\bottomrule
& LOCUS 2.0  & $2.42$ & $3.88$ & $107.72$ & $185.00$ & $2.13$\\ 
J & FAST-LIO & $\textbf{1.72}$ & $\textbf{2.60}$ & $126.72$ & $332.50$ & $2.54$\\
& LINS & $3.56$ & $5.79$ & $\textbf{73.76}$ & $\textbf{176.50}$ & $\textbf{1.85}$ \\
\bottomrule
        \end{tabular}
		}
	\end{center}
\end{table}
}

\section{Conclusions}
This work presents LOCUS 2.0, a robust and computational efficient \lidar odometry system for real-time, large-scale explorations under severe computation and memory constraints suitable to be deployed over heterogeneous robotic platforms. 
This work reformulates GICP covariance calculations from precomputed normals that improves the computational performance of GICP.
LOCUS 2.0 uses an adaptive voxel grid filter and makes the computational load independent on the environment and sensor configuration.
Adaptive behavior keeps the number of points from the \lidar consistent while keeping the voxelized structure of the environment, which stabilizes and improves the computational load.
\morrell{We evaluate two sliding map strategies for reducing memory use: multi-threaded octree and \ikdtree, and show both their computational cost, and improvement in memory usage. We open-source both LOCUS 2.0, and our dataset for challenging and large-scale underground environments that features various real-world conditions such as fog, dust, darkness, and geometrically degenerate environments that restrict mobility. Overall the datasets include  $11~h$ of operations and $16~km$ distance traveled.}


%





\ifCLASSOPTIONcaptionsoff
  \newpage
\fi

\vspace{-0.5em}
\bibliographystyle{IEEEtran}
\bibliography{IEEEexample}








\end{document}